\documentclass[10pt,twocolumn,letterpaper]{article}

\usepackage[pagenumbers]{iccv} 

\definecolor{iccvblue}{rgb}{0.21,0.49,0.74}
\usepackage[pagebackref,breaklinks,colorlinks,allcolors=iccvblue]{hyperref}

\usepackage{caption}
\usepackage{subcaption}
\usepackage{xcolor}
\usepackage{esvect}
\usepackage{amsmath}
\usepackage{bm}
\usepackage{dsfont}
\usepackage{svg}
\usepackage{float}
\usepackage{array}

\newcommand{\PreserveBackslash}[1]{\let\temp=\\#1\let\\=\temp}
\newcolumntype{C}[1]{>{\PreserveBackslash\centering}p{#1}}
\newcolumntype{R}[1]{>{\PreserveBackslash\raggedleft}p{#1}}
\newcolumntype{L}[1]{>{\PreserveBackslash\raggedright}p{#1}}

\def\lc{\lfloor}   
\def\rc{\rfloor}

\newcommand{\AB}[1]{\textcolor{black}{#1}}

\definecolor{tabblu}{HTML}{1f77b4}
\definecolor{taborg}{HTML}{ff7f0e}
\definecolor{tabgre}{HTML}{2ca02c}
\definecolor{tabred}{HTML}{d62728}
\definecolor{tabpur}{HTML}{9467bd}
\definecolor{tabbrw}{HTML}{8c564b}
\definecolor{tabpnk}{HTML}{e377c2}
\definecolor{tabgry}{HTML}{7f7f7f}
\definecolor{tabolv}{HTML}{bcbd22}
\definecolor{tabcyn}{HTML}{17becf}

\definecolor{magenta}{HTML}{dc143c}
\definecolor{purple}{HTML}{4b0082}

\usepackage{xspace}
\usepackage{tikz}
\usepackage{listings}
\definecolor{codegreen}{rgb}{0,0.3,0}
\definecolor{codegray}{rgb}{0.5,0.5,0.5}
\definecolor{codepurple}{rgb}{0.58,0,0.82}
\definecolor{backcolour}{rgb}{0.95,0.95,0.92}

\lstdefinestyle{mystyle}{
    backgroundcolor=\color{backcolour},   
    commentstyle=\color{codegreen},
    keywordstyle=\color{magenta},
    numberstyle=\tiny\color{codegray},
    stringstyle=\color{codepurple},
    basicstyle=\ttfamily\footnotesize,
    breakatwhitespace=false,         
    breaklines=true,                 
    captionpos=b,                    
    keepspaces=true,                 
    numbers=left,                    
    numbersep=5pt,                  
    showspaces=false,                
    showstringspaces=false,
    showtabs=false,                  
    tabsize=2
}

\newcommand{\ILoRA}{\texorpdfstring{$\mathds{1}$}{1}LoRA\xspace}

\def\radius{2pt}

\newcommand*{\all}{%
  \begin{tikzpicture}
    \node[circle,fill=tabblu,inner sep=\radius] at (0,0) {};
  \end{tikzpicture}%
  \xspace%
}

\newcommand*{\lora}{%
  \begin{tikzpicture}
    \node[circle,fill=tabbrw,inner sep=\radius] at (0,0) {};
  \end{tikzpicture}%
  \xspace%
}

\newcommand*{\loraqkv}{%
  \begin{tikzpicture}
    \node[circle,fill=tabcyn,inner sep=\radius] at (0,0) {};
  \end{tikzpicture}%
  \xspace%
}

\newcommand*{\ilora}{%
  \begin{tikzpicture}
    \node[circle,fill=tabolv,inner sep=\radius] at (0,0) {};
  \end{tikzpicture}%
  \xspace%
}

\newcommand*{\mora}{%
  \begin{tikzpicture}
    \draw[color=taborg,fill=taborg,line width=0] (0,.098) -- (0,.098) arc(270:90:-.098) --cycle;
    \draw[color=tabgre,fill=tabgre,line width=0] (0,.098) -- (0,.098) arc(90:270: .098) --cycle;
  \end{tikzpicture}%
  \xspace%
}

\newcommand*{\morasix}{%
  \begin{tikzpicture}
    \node[circle,fill=tabgre,inner sep=\radius] at (0,0) {};
  \end{tikzpicture}%
  \xspace%
}

\newcommand*{\bitfit}{%
  \begin{tikzpicture}
    \node[circle,fill=tabred,inner sep=\radius] at (0,0) {};
  \end{tikzpicture}%
  \xspace%
}

\newcommand*{\difffit}{%
  \begin{tikzpicture}
    \node[circle,fill=tabpur,inner sep=\radius] at (0,0) {};
  \end{tikzpicture}%
  \xspace%
}

\newcommand*{\vera}{%
  \begin{tikzpicture}
    \node[circle,fill=tabpnk,inner sep=\radius] at (0,0) {};
  \end{tikzpicture}%
  \xspace%
}

\newcommand*{\dora}{%
  \begin{tikzpicture}
    \node[circle,fill=tabgry,inner sep=\radius] at (0,0) {};
  \end{tikzpicture}%
  \xspace%
}

\newcommand*{\iloranorms}{%
  \begin{tikzpicture}
    \node[circle,fill=tabpnk,inner sep=\radius] at (0,0) {};
  \end{tikzpicture}%
  \xspace%
}

\newcommand*{\iloranormsFigClassification}{%
  \begin{tikzpicture}
    \node[circle,fill=tabcyn,inner sep=\radius] at (0,0) {};
  \end{tikzpicture}%
  \xspace%
}

\newcommand*{\ilorabias}{%
  \begin{tikzpicture}
    \node[circle,fill=tabbrw,inner sep=\radius] at (0,0) {};
  \end{tikzpicture}%
  \xspace%
}

\newcommand*{\ilorac}{%
  \begin{tikzpicture}
    \node[circle,fill=tabpur,inner sep=\radius] at (0,0) {};
  \end{tikzpicture}%
  \xspace%
}

\newcommand*{\norms}{%
  \begin{tikzpicture}
    \node[circle,fill=tabblu,inner sep=\radius] at (0,0) {};
  \end{tikzpicture}%
  \xspace%
}

\newcommand*{\bias}{%
  \begin{tikzpicture}
    \node[circle,fill=taborg,inner sep=\radius] at (0,0) {};
  \end{tikzpicture}%
  \xspace%
}

\newcommand*{\scaling}{%
  \begin{tikzpicture}
    \node[circle,fill=tabgre,inner sep=\radius] at (0,0) {};
  \end{tikzpicture}%
  \xspace%
}

\usepackage[most]{tcolorbox} 
\usepackage{xcolor}          
\usepackage{multicol}        
\usepackage{multirow}        
\usepackage{parskip}         

\tcbset{
  question/.style={
    enhanced,
    colback=black!5,          
    colframe=black!50,        
    sharp corners=southwest, 
    boxrule=0.5mm,           
    width=\dimexpr0.5\textwidth-1em\relax, 
    left=1em,
    right=1em,
    top=0.5em,
    bottom=0.5em,
    fonttitle=\bfseries,     
    title=Question \& Answer          
  },
  ilora/.style={
    enhanced,
    colback=tabolv!5,          
    colframe=tabolv!50,        
    sharp corners=southeast, 
    boxrule=0.5mm,           
    width=\dimexpr0.5\textwidth-1em\relax, 
    left=1em,
    right=1em,
    top=0.5em,
    bottom=0.5em,
    fonttitle=\bfseries,     
    title=\ILoRA          
  },
  loraqkv/.style={
    enhanced,
    colback=tabcyn!5,         
    colframe=tabcyn!50,       
    sharp corners=southeast, 
    boxrule=0.5mm,
    width=\dimexpr0.5\textwidth-1em\relax, 
    left=1em,
    right=1em,
    top=0.5em,
    bottom=0.5em,
    fonttitle=\bfseries,     
    title=LoRA QKV          
  }
}

\def\paperID{11188} 
\def\confName{ICCV}
\def\confYear{2025}

\title{\ILoRA: Summation Compression for Very Low-Rank Adaptation}

\author{\textbf{Alessio Quercia} $^{1,3}$ \quad
\textbf{Zhuo Cao} $^{1}$ \quad
\textbf{Arya Bangun} $^{1}$ \quad
\textbf{Richard D.~Paul} $^{1,4}$ \\
\textbf{Abigail Morrison} $^{2,3}$ \quad
\textbf{Ira Assent} $^{1,5}$ \quad
\textbf{Hanno Scharr} $^{1}$ \\ \newline
{$^{1}$ IAS-8 $^{2}$ IAS-6, Forschungszentrum Juelich, Juelich, Germany } \\
{$^{3}$ Dept. of Computer Science, RWTH Aachen University, Aachen, Germany} \\
{$^{4}$ Dept. of Statistics, LMU Munich, Munich, Germany} \\ 
{$^{5}$ Dept. of Computer Science, Aarhus University, Aarhus, Denmark} \\ 
{\tt\small\{a.quercia,z.cao,a.bangun,r.paul,a.morrison,i.assent,h.scharr\}@fz-juelich.de}
}

\usepackage{tikz}
\newcommand{\slant}[2]{%
  \tikz[baseline=(X.base), xslant=tan(#1)]
  \node[inner sep=0pt, outer sep=0pt, xslant=tan(#1)](X){#2};%
  \!%
}

\begin{document}
\maketitle

\begin{abstract}
    Parameter-Efficient Fine-Tuning (PEFT) methods have transformed the approach to fine-tuning large models for downstream tasks by enabling the adjustment of significantly fewer parameters than those in the original model matrices. In this work, we study the "very low rank regime", where we fine-tune the lowest amount of parameters per linear layer for each considered PEFT method. We propose \slant{15}{$\mathds{1}$}LoRA 
    (Summation Low-Rank Adaptation), 
    a compute, parameter and memory efficient fine-tuning method which uses the feature sum as fixed compression and a single trainable vector as decompression. Differently from state-of-the-art PEFT methods like LoRA, VeRA, and the recent MoRA, \slant{15}{$\mathds{1}$}LoRA uses fewer parameters per layer, reducing the memory footprint and the computational cost. We extensively evaluate our method against state-of-the-art PEFT methods on multiple fine-tuning tasks, and show that our method not only outperforms them, but is also more parameter, memory and computationally efficient. Moreover, thanks to its memory efficiency, \slant{15}{$\mathds{1}$}LoRA allows to fine-tune more evenly across layers, instead of focusing on specific ones (\eg attention layers), improving performance further.
\end{abstract}

\section{Introduction}
In recent years, the rapid development of large models has transformed multiple fields in artificial intelligence, especially in natural language processing \cite{achiam2023gpt,touvron2023llama} and computer vision \cite{kirillov2023segment,depthanything,dit}. However, adapting these models for specific tasks often demands significant computational power and extensive labeled data, making their deployment challenging for many applications. To mitigate these issues, Parameter Efficient Fine-Tuning (PEFT) methods \cite{zaken2021bitfit,xie2023difffit,hu2021lora,liu2024dora,jiang2024mora,kopiczko2023vera} have emerged as a promising approach to modify pre-trained models with minimal adjustments to their parameters, thereby reducing computational and memory demands while preserving performance.

\begin{figure}[t]
     \centering
     \begin{subfigure}[h]{0.2\textwidth}
         \centering
         \resizebox{\columnwidth}{!}{
\begingroup%
  \makeatletter%
  \providecommand\color[2][]{%
    \errmessage{(Inkscape) Color is used for the text in Inkscape, but the package 'color.sty' is not loaded}%
    \renewcommand\color[2][]{}%
  }%
  \providecommand\transparent[1]{%
    \errmessage{(Inkscape) Transparency is used (non-zero) for the text in Inkscape, but the package 'transparent.sty' is not loaded}%
    \renewcommand\transparent[1]{}%
  }%
  \providecommand\rotatebox[2]{#2}%
  \newcommand*\fsize{\dimexpr\f@size pt\relax}%
  \newcommand*\lineheight[1]{\fontsize{\fsize}{#1\fsize}\selectfont}%
  \ifx\svgwidth\undefined%
    \setlength{\unitlength}{149.04576665bp}%
    \ifx\svgscale\undefined%
      \relax%
    \else%
      \setlength{\unitlength}{\unitlength * \real{\svgscale}}%
    \fi%
  \else%
    \setlength{\unitlength}{\svgwidth}%
  \fi%
  \global\let\svgwidth\undefined%
  \global\let\svgscale\undefined%
  \makeatother%
  \begin{picture}(1,1.54410806)%
    \lineheight{1}%
    \setlength\tabcolsep{0pt}%
    \put(0,0){\includegraphics[width=\unitlength,page=1]{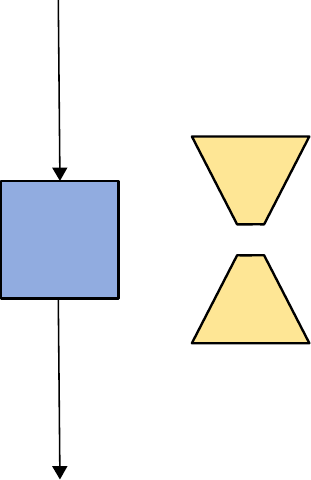}}%
    \put(0.1246517,0.74615042){\makebox(0,0)[lt]{\lineheight{1.25}\smash{\begin{tabular}[t]{l}$\bm{W_0}$\end{tabular}}}}%
    \put(0.77170011,0.93436853){\makebox(0,0)[lt]{\lineheight{1.25}\smash{\begin{tabular}[t]{l}$\bm{A}$\end{tabular}}}}%
    \put(0.76980786,0.52028){\makebox(0,0)[lt]{\lineheight{1.25}\smash{\begin{tabular}[t]{l}$\bm{B}$\end{tabular}}}}%
    \put(0,0){\includegraphics[width=\unitlength,page=2]{diagram_lora_svg-tex.pdf}}%
  \end{picture}%
\endgroup%
}
         \caption{LoRA}
         \label{fig:lora}
     \end{subfigure}
     \hspace{1cm}
     \begin{subfigure}[h]{0.2\textwidth}
         \centering
         \resizebox{\columnwidth}{!}{
\begingroup%
  \makeatletter%
  \providecommand\color[2][]{%
    \errmessage{(Inkscape) Color is used for the text in Inkscape, but the package 'color.sty' is not loaded}%
    \renewcommand\color[2][]{}%
  }%
  \providecommand\transparent[1]{%
    \errmessage{(Inkscape) Transparency is used (non-zero) for the text in Inkscape, but the package 'transparent.sty' is not loaded}%
    \renewcommand\transparent[1]{}%
  }%
  \providecommand\rotatebox[2]{#2}%
  \newcommand*\fsize{\dimexpr\f@size pt\relax}%
  \newcommand*\lineheight[1]{\fontsize{\fsize}{#1\fsize}\selectfont}%
  \ifx\svgwidth\undefined%
    \setlength{\unitlength}{149.04581447bp}%
    \ifx\svgscale\undefined%
      \relax%
    \else%
      \setlength{\unitlength}{\unitlength * \real{\svgscale}}%
    \fi%
  \else%
    \setlength{\unitlength}{\svgwidth}%
  \fi%
  \global\let\svgwidth\undefined%
  \global\let\svgscale\undefined%
  \makeatother%
  \begin{picture}(1,1.54410747)%
    \lineheight{1}%
    \setlength\tabcolsep{0pt}%
    \put(0,0){\includegraphics[width=\unitlength,page=1]{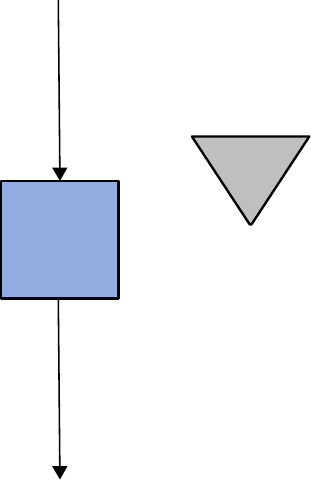}}%
    \put(0.12465159,0.74615014){\makebox(0,0)[lt]{\lineheight{1.25}\smash{\begin{tabular}[t]{l}$\bm{W_0}$\end{tabular}}}}%
    \put(0,0){\includegraphics[width=\unitlength,page=2]{diagram_ilora_svg-tex.pdf}}%
    \put(0.88317047,0.52153303){\makebox(0,0)[lt]{\lineheight{1.25}\smash{\begin{tabular}[t]{l}$\bm{b}$\end{tabular}}}}%
    \put(0.75004013,0.95896129){\makebox(0,0)[lt]{\lineheight{1.25}\smash{\begin{tabular}[t]{l}$\bm{\sum}$\end{tabular}}}}%
  \end{picture}%
\endgroup%
}
         \caption{\ILoRA}
         \label{fig:ilora}
     \end{subfigure}
    \caption{Comparing our method \ILoRA to LoRA. Left: LoRA learns the low-rank decomposition $\Delta W = BA$, where $A\in \mathbb{R}^{r \times k}$ and $B\in \mathbb{R}^{d \times r}$. Right: \ILoRA replaces the matrices A and B with a sum over the input features $x$ 
    as compression and a learnable vector $b\in \mathbb{R}^{1 \times d}$ as decompression: $\Delta W = b\mathds{1}^T$, where the feature sum is $\mathds{1}^Tx = \sum_{i=1}^k x_i$, with $\mathds{1}$ being a vector of length $k$ containing only ones. This reduces the trainable parameters from $r \times k + d \times r$ in LoRA to $d$ per layer for \ILoRA.}
    \label{fig:diagram}
\end{figure}

Among various PEFT approaches, BitFit \cite{zaken2021bitfit} has gained notable attention. It streamlines fine-tuning by updating only the bias parameters in transformer models, offering a lightweight alternative achieving competitive performance with minimal resource consumption. DiffFit \cite{xie2023difffit} 
extends this method
by adding a scaling factor to linear layers and also updating normalization layers, enabling rapid adaptation of large pre-trained diffusion models to new domains. These methods significantly speed up training and reduce model storage requirements, allowing practitioners to fine-tune models effectively without extensive retraining of all parameters. Yet, because weight matrices are not changed
%
%
they may not reach the performance achieved by other methods, such as Low-Rank Adaptation (LoRA) \cite{hu2021lora, pu2023empirical}.

LoRA \cite{hu2021lora} and its variants \cite{kopiczko2023vera,jiang2024mora,liu2024dora,dettmers2024qlora,guo2023lq,li2024vb} introduce low-rank matrices for model weight updates during fine-tuning, enabling efficient training with fewer parameters, see \Cref{fig:diagram} (left). This approach not only enhances the adaptability of large models but also reduces training time and mitigates overfitting. Compared to BitFit, LoRA achieves stronger performance in certain tasks and demonstrates robust generalization capabilities \cite{xin-etal-2024-beyond, razuvayevskaya2024comparison}. This performance boost, however, requires more trainable parameters, which has been shown to cause training instability and slower convergence \cite{chen2022revisiting}. Additionally, Pu et al.\ \cite{pu2023empirical} observe that LoRA is robust to parameter reduction for text classification and generation tasks, allowing for reductions beyond the original proposed ones. This raises the question: can we maintain or improve performance while reducing LoRA’s trainable parameters to levels comparable to BitFit?

In this work, we propose \ILoRA, a method centered on very low-rank adaptation, as depicted in \Cref{fig:diagram}. We reduce the parameter count to match that of BitFit \cite{zaken2021bitfit}, making it significantly lower than that required by other low-rank adaptation methods. \ILoRA applies a predefined non-trainable function for input compression, \ie\ the sum over all input features $x$
and utilizes a single trainable vector per layer for decompression. As we demonstrate in our extensive, comparative experiments \ILoRA provides enhanced performance and memory efficiency in different tasks.

Our contributions can be summarized as follows:
\begin{itemize}
    \item We propose \ILoRA, a very low-rank adaptation method 
    which fixes the compression mechanism to a simple feature sum
    and a vector of $d$ parameters for decompression, where $d$ is the number of output features, significantly reducing the number of trainable parameters.
    \item Due to its low memory footprint at training time, \ILoRA enables fine-tuning across more layers in larger models (\eg, LLaMA2 7 and 13 billion in our experiments), distributing adjustments across multiple layers instead of focusing solely on limited ones (\eg\ the attention layers), thereby achieving enhanced performance.
    \item \ILoRA adds no overhead at inference time, as the update $\Delta W$ can be merged with the initial weight matrix $W_0$ in the same way as \eg\ done in LoRA \cite{hu2021lora}.
    \item Our method is simple to implement, making it accessible for practical deployment.
    \item We conduct extensive empirical evaluations demonstrating that \ILoRA outperforms state-of-the-art methods in terms of effectiveness, and parameter, memory and computational efficiencies across multiple tasks and datasets.
\end{itemize}

\section{Related Work} \label{sec: related_work}
PEFT methods aim to adapt large pre-trained models for specific downstream tasks by injecting a small number of learnable parameters while freezing the majority of the model's parameters. This approach addresses the resource inefficiency associated with full fine-tuning methods, where all model parameters are tuned for the new task. 

Various strategies are employed to integrate learnable parameters in PEFT. For instance, Adapter \cite{houlsby2019parameter} methods introduce small, fine-tunable modules across different layers of Transformer models, optimizing only these new modules for downstream adaptation. Prompt-tuning \cite{lester2021power} and Prefix-tuning \cite{li2021prefix} involve inserting a set of trainable tokens into the input layer of Transformers, focusing training efforts on these tokens only. More recently, LoRA \cite{hu2021lora} reparameterizes certain pre-trained weights using low-rank matrices, tuning only these additional branches to adapt to new tasks. BitFit \cite{zaken2021bitfit} proposes to fine-tune only bias terms, minimizing the number of trainable parameters. Although updating only a fraction of the parameters, these PEFT methods achieve comparable or even superior performance to full fine-tuning approaches.

While PEFT methods were initially developed within the natural language processing domain, recent works have extended their application to computer vision tasks. For example, \cite{zhong2024convolution} apply LoRA on the Segment Anything model \cite{kirillov2023segment}, and DiffFit \cite{xie2023difffit} transfers BitFit to fine-tune the diffusion model. Moreover, PEFT methods have recently gained interest for subspace Bayesian inference in large-scale Transformers \cite{onal2024gaussian, yang2024bayesian, paul2024parameterefficientbayesianneuralnetworks},demonstrating the effectiveness of this efficient tuning paradigm across multiple domains.

\section{Very Low-Rank Adaptation}\label{sec: method}
In this section, we first analyze existing PEFT methods in regards to their fine-tuning approach and number of trainable parameters. We then motivate and describe technical details of our memory efficient very-low rank adaptation method, \ILoRA. Finally, we compare our method with other methods in terms of the number of parameters and computational and memory efficiency, respectively. 

\subsection{PEFT methods}  \label{sec: preliminary}
We start by reviewing the technical details of state-of-the-art PEFT methods, analyzing in particular their computational and memory requirements. We provide an overview for easy comparison in \Cref{tab:baselines}.

\begin{table}[tb]
\setlength{\tabcolsep}{5pt}
\centering
\caption{Analysed PEFT methods: number of trainable parameters in the "very low-rank regime" with $r=1$ (LoRA, VeRA and DoRA) and  $\hat{r}=\lc \sqrt{d} \rc$ (MoRA). 
}
\begin{tabular}{ L{2.15cm} C{3.1cm} C{2.cm} }
    \toprule
    Method & Equation & \#Params $\times$ Linear \\
    \midrule

    LoRA \cite{hu2021lora} & $W_0 x + \underline{BA} x$ & $k + d$ \\[2mm]
    DoRA \cite{liu2024dora} & $W_0 x + \underline{m}\frac{W_0 + \underline{BA}}{||W_0 + \underline{BA}||_c} x$ & $d + k + d$ \\[2mm]
    VeRA \cite{kopiczko2023vera} & $W_0 x + \underline{b}B\underline{d}A x$ & $1 + d$ \\[2mm]
    MoRA$_{1, 6}$ \cite{jiang2024mora} & $W_0 x + f_d(\underline{M}f_c(x))$ & $\hat{r}^2$ \\[2mm] 
    BitFit \cite{zaken2021bitfit} & $W_0 x + \underline{\beta}$ & $d$ \\[2mm]
    DiffFit \cite{xie2023difffit} & $\underline{\gamma}(W_0 x + \underline{\beta})$, \underline{norms} & $>2d$ \\[2mm]
    \ILoRA & $W_0 x + \underline{b} \mathds{1}^Tx$ & $d$ \\
    \midrule
    All & $\underline{W_0} x + \underline{\beta}$, \underline{norms}  & $>k \times d + d$ \\
    \bottomrule
\end{tabular}
\label{tab:baselines}
\end{table}

\paragraph{LoRA} \cite{hu2021lora} was one of the first methods proposed for parameter efficient fine-tuning. Instead of fine-tuning the entire parameter matrix $W_0 \in \mathbb{R}^{d\times k}$, LoRA fine-tunes an additive shift matrix $\Delta W \in \mathbb{R}^{d \times k}$ as follows:
\begin{equation} \label{eq:shift}
    o = W_0 x + \underline{\Delta W} x = (W_0 + \underline{\Delta W}) x
\end{equation}
where $x \in \mathbb{R}^{k}$ and $o \in \mathbb{R}^{d}$ are input and output features, respectively. We \underline{underline} symbols indicating trainable parameters to distinguish them from pretrained, frozen parameters. LoRA decomposes the shift matrix ${\Delta W}$ by two rectangular matrices:
\begin{align}
    \Delta W = \underline{BA}, \quad \text{with} \ A \in \mathbb{R}^{r \times k}, B \in \mathbb{R}^{d \times r}.
\end{align}
Matrix $A$ reduces the rank of input from $k$ to $r$ and matrix $B$ decompresses it to the size of output $d$. 

\vspace{-3mm}
\paragraph{DoRA} (weight-decomposed low-rank adaptation) \cite{liu2024dora} builds on LoRA and proposes to decompose the pretrained matrices into magnitude $m \in \mathbb{R}^{d}$ and direction $({W_0 + \underline{BA}})/({||W_0 + \underline{BA}||_c})$. Here, the norm is defined as vector-wise norm of a matrix across each column.

\vspace{-3mm}
\paragraph{VeRA} (Vector-based Random-matrix Adaptation) \cite{kopiczko2023vera} builds on top of LoRA using fixed random compression and decompression matrices ($A$ and $B$), where the only trainable parameters are two vectors ($d \in \mathbb{R}^{r}$ and $b \in \mathbb{R}^{d}$), respectively after the compression and decompression. 

\vspace{-3mm}
\paragraph{MoRA} \cite{jiang2024mora}, similar to  {VeRA}, also uses fixed compression $f_c$ and decompression $f_d$ schemes, while only learning the inner low-rank squared matrix $M \in \mathbb{R}^{\hat{r} \times \hat{r}}$, where the rank is $\hat{r} = \lc \sqrt{(k + d)r} \rc$.
The authors of \cite{jiang2024mora} propose two versions, named types 1 and 6 in their Python code\footnote{\url{https://github.com/kongds/MoRA/}}. The version 'type 1 (Sharing)' is according to Eq.~6 in \cite{jiang2024mora} and uses a sum within groups of features as compression and the decompression is a copy of learned features. The 'type 6 (RoPE based)' is a version according to Eq.~9 in \cite{jiang2024mora} where the features are reshaped into a new axis as compression, processed, and then reshaped back. We name these methods MoRA$_1$ and MoRA$_6$, respectively. 

\vspace{-3mm}
\paragraph{BitFit} \cite{zaken2021bitfit} is one of the first PEFT methods, proposing to fine-tune only the biases $\beta \in \mathbb{R}^d$ of a pre-trained model. 

\vspace{-3mm}
\paragraph{DiffFit} \cite{xie2023difffit} extends BitFit to diffusion models by introducing a scaling vector $\gamma \in \mathbb{R}^{d}$. \AB{As a result, this approach adds an additional parameter that requires fine-tuning.} Moreover, DiffFit trains the normalization layers and the class embedding, if present in the model.


\vspace{-3mm}
\bigskip
\noindent
In this paper, we explore the "very low-rank regime", where we use the smallest amount of parameters for each state-of-the-art method. We notice that BitFit \cite{zaken2021bitfit} and VeRA \cite{kopiczko2023vera} are the methods that use the least parameters in this scenario, with $d$ and $1 + d$ per layer, when the rank is $r = 1$. All other methods use more parameters. VeRA smartly shares the constant random matrices A and B across modules with the same shape to reduce computation, however, it still uses considerable memory for matrix storage. BitFit, on the other hand, only fine-tunes the bias terms that remain constant across inputs and therefore lack generalization capabilities. Furthermore, we notice that MoRA \cite{jiang2024mora} has memory and computational inefficiencies due to the grouping and duplication operations, whereas other state-of-the-art LoRA methods heavily rely on the decomposition matrices A and B, which requires them to use additional memory. 

\subsection{\ILoRA}
To reduce the memory consumption and to allow for better scaling to big models, we adopt a very low rank regime. Concretely, we propose to compress the input features to their sum $\sum_{i=1}^k x_i$. This sum can be expressed as $\mathds{1}^Tx$, where $\mathds{1}$ is the main diagonal of the all-positive quadrant. Our intuition is based on the fact that typical non-linearities (\eg\ ReLU) move features towards this quadrant. Thus $\mathds{1}$ is a good candidate for a vector strongly correlating with the inputs when we have this prior knowledge, and it is as good as any other random vector when we do not. 

Our approach can be viewed an extreme case of MoRA \cite{jiang2024mora} which compresses the input features by computing the sum within specific groups of features and decompresses them by duplicating their outputs. 
Importantly, in contrast to MoRA, \ILoRA uses a single trainable vector $b$ of $d$ parameters for decompression, thereby retaining performance in fine-tuning. Using vector $\mathds{1}$ of length $k$ containing only ones, the shift matrix can be written as
\begin{align}
\label{eq:ilora}
    \Delta W &= \underline{b} \mathds{1}^T & \mathds{1} \in \{1\}^{k \times 1}, b \in \mathbb{R}^{d \times 1}
\end{align} 
\Cref{fig:diagram} depicts \ILoRA in comparison to LoRA \cite{hu2021lora}. 
%
%
In practice, \ILoRA only introduces a trainable vector of $d$ parameters per layer. The compression can be implemented in PyTorch \cite{pytorch} as a sum over the features, removing the need for additional compression and decompression matrices, as in most state-of-the-art methods \cite{hu2021lora,kopiczko2023vera,jiang2024mora,liu2024dora}. This makes our method more computationally and memory efficient. 

For fast and memory efficient inference, 
weights can be merged as the right side in \Cref{eq:shift}, \ie\ the fine-tuned weight matrix ${W}_\text{ft}$ is given as ${W}_\text{ft}:=W_0+\Delta W$. In our experiments, this efficient implementation has been applied to the competitors LoRA, MoRA, and DoRA \cite{hu2021lora,jiang2024mora, liu2024dora}.

\subsection{\ILoRA Method Analysis} 
Our method relates to LoRA \cite{hu2021lora} with rank $r=1$ which fine-tunes the low-rank matrices A and B. Compared to LoRA, \ILoRA uses $k$ fewer parameters per layer (see \Cref{tab:baselines}), while removing the need for the matrices A and B. 
MoRA (Eq. 6, \cite{jiang2024mora}) introduces constant, non-trainable compression and decompression functions: a sum over feature groups for compression and a duplication of learned features for decompression. Our approach goes to the most extreme case, grouping features by summing across a single group that includes all features. Simultaneously, it eliminates the need for copying learned features by introducing the vector $b$, which scales the feature sum into $d$ parameters during decompression. Although MoRA can be made comparable to \ILoRA in terms of parameters by using rank $\hat{r}=\lc \sqrt{d} \rc$, \ILoRA is more computationally efficient as it avoids the need for grouping features, learning the inner features, and copying them back to match the required shape. 

Among low-rank adaptation methods, only VeRA has a comparable number of parameters. However, it uses random projection matrices, making the method more memory-hungry, initialization-dependent, and less interpretable. By contrast, using the feature sum as input for the trainable decompression, \ILoRA suggests that it suffices to learn to shift the pre-trained parameters by scaled feature sum of samples drawn from a dataset to learn the new task.

In comparison to BitFit, which only fine-tunes the biases $\beta$, \ILoRA can better learn from the data as it uses the sum of the features, resulting in better performance, while slightly sacrificing training speed, see experiments in \Cref{sec:Experiments}. Lastly, our method is complementary to state-of-the-art methods that fine-tune additional terms, like biases (BitFit \cite{zaken2021bitfit}) or normalization layers (DiffFit, \cite{xie2023difffit}), and to methods building on top of LoRA \cite{hu2021lora,liu2024dora,dettmers2024qlora,guo2023lq,li2024vb}.


\section{Empirical evaluation}\label{sec:Experiments}
We benchmark \ILoRA against state-of-the-art PEFT methods in the "very low-rank regime" on multiple fine-tuning tasks and models:  Monocular Depth Estimation with DepthAnything \cite{depthanything}, Image Classification with ViT-Base \cite{vit}, Mathematical Reasoning with LLaMA2 7b and 13b \cite{llama2}, and Image Generation with DiT-XL-2-256x256 \cite{dit}. For each method we use the lowest amount of parameters per layer. We compare \ILoRA with LoRA ($r=1$), VeRA ($r=1$), DoRA ($r=1$), MoRA ($\hat{r}=\lc \sqrt{d} \rc$) types~1 and~6 (see \Cref{sec: preliminary}), BitFit and DiffFit. We report results in forms of tables and bubble plots, where the mean of 4 different independent runs is reported for each method. In each plot, we report the best mean evaluation metrics as y-axis, and their required total (training and validation) wall clock time as x-axis, GPU (A100 40GB) memory consumption as bar plot, and number of trainable parameters as bubble area. Note that parameters are only a proxy for memory consumption, which can be higher depending on computational graph complexity. We report metrics' standard deviations in the table and plots (as error bars).

\begin{table}[tb]
\setlength{\tabcolsep}{1pt}
\centering
\caption{RMSE ($\downarrow$) and AbsRel ($\downarrow$) of DepthAnything model pre-trained on KITTI and fine-tuned to NYU using PEFT methods. LoRA, VeRA and DoRA with rank $r=1$, MoRA with rank $\hat{r}=\lc \sqrt{d} \rc$  and $\tilde{d} = \hat{r}^2$ parameters.}
\resizebox{\columnwidth}{!}{
\begin{tabular}{ lcccc} 
    \toprule
    \multirow{2}{*}{Method} & \multicolumn{2}{c}{KITTI $\to$ NYU} & \multicolumn{2}{c}{NYU $\to$ KITTI} \\
                & RMSE ($\downarrow$)           & AbsRel ($\downarrow$)          & RMSE ($\downarrow$)           & AbsRel ($\downarrow$)            \\
    \midrule 
    MoRA$_1$    & $0.248    _{_{\pm 0.00066}}$  & $0.0670    _{_{\pm 0.00018}}$  & $2.509    _{_{\pm 0.00455}}$  & $0.0587       _{_{\pm 0.00011}}$ \\ 
    MoRA$_6$    & $0.277    _{_{\pm 0.00021}}$  & $0.0774    _{_{\pm 0.00009}}$  & $2.971    _{_{\pm 0.00239}}$  & $0.0739       _{_{\pm 0.00009}}$ \\ 
    LoRA        & $0.250    _{_{\pm 0.00087}}$  & $0.0678    _{_{\pm 0.00028}}$  & $2.353    _{_{\pm 0.00315}}$  & $0.0583       _{_{\pm 0.00010}}$ \\
    VeRA        & $0.244    _{_{\pm 0.00059}}$  & $0.0668    _{_{\pm 0.00013}}$  & $2.304    _{_{\pm 0.00524}}$  & $0.0577       _{_{\pm 0.00013}}$ \\
    DoRA        & $0.249    _{_{\pm 0.00088}}$  & $0.0675    _{_{\pm 0.00031}}$  & $2.345    _{_{\pm 0.00275}}$  & $0.0581       _{_{\pm 0.00004}}$ \\
    BitFit      & $0.349    _{_{\pm 0.00073}}$  & $0.1015    _{_{\pm 0.00022}}$  & $4.806    _{_{\pm 0.00392}}$  & $0.1271       _{_{\pm 0.00027}}$ \\
    DiffFit     & $0.314    _{_{\pm 0.00078}}$  & $0.0895    _{_{\pm 0.00030}}$  & $3.904    _{_{\pm 0.00539}}$  & $0.0988       _{_{\pm 0.00024}}$ \\
    \ILoRA       & $\bm{0.238_{_{\pm 0.00050}}}$ & $\bm{0.0647_{_{\pm 0.00010}}}$ & $\bm{2.203_{_{\pm 0.00780}}}$ & $\bm{0.0533   _{_{\pm 0.00025}}}$\\ \midrule
    All         & $0.210    _{_{\pm 0.00064}}$  & $0.0571    _{_{\pm 0.00033}}$  & $1.916    _{_{\pm 0.00303}}$  & $0.0463       _{_{\pm 0.00010}}$ \\
    \bottomrule
\end{tabular}
}
\label{tab:nyu_to_kitti}
\end{table}

\begin{figure*}[tb]
     \centering
     \begin{subfigure}[h]{\textwidth}
         \centering
         \resizebox{0.4\columnwidth}{!}{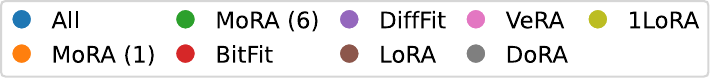}
         \vspace{0.3cm}
     \end{subfigure}
     \vspace{0.3cm}
     \begin{subfigure}[h]{0.49\textwidth}
         \centering
         \resizebox{\columnwidth}{!}{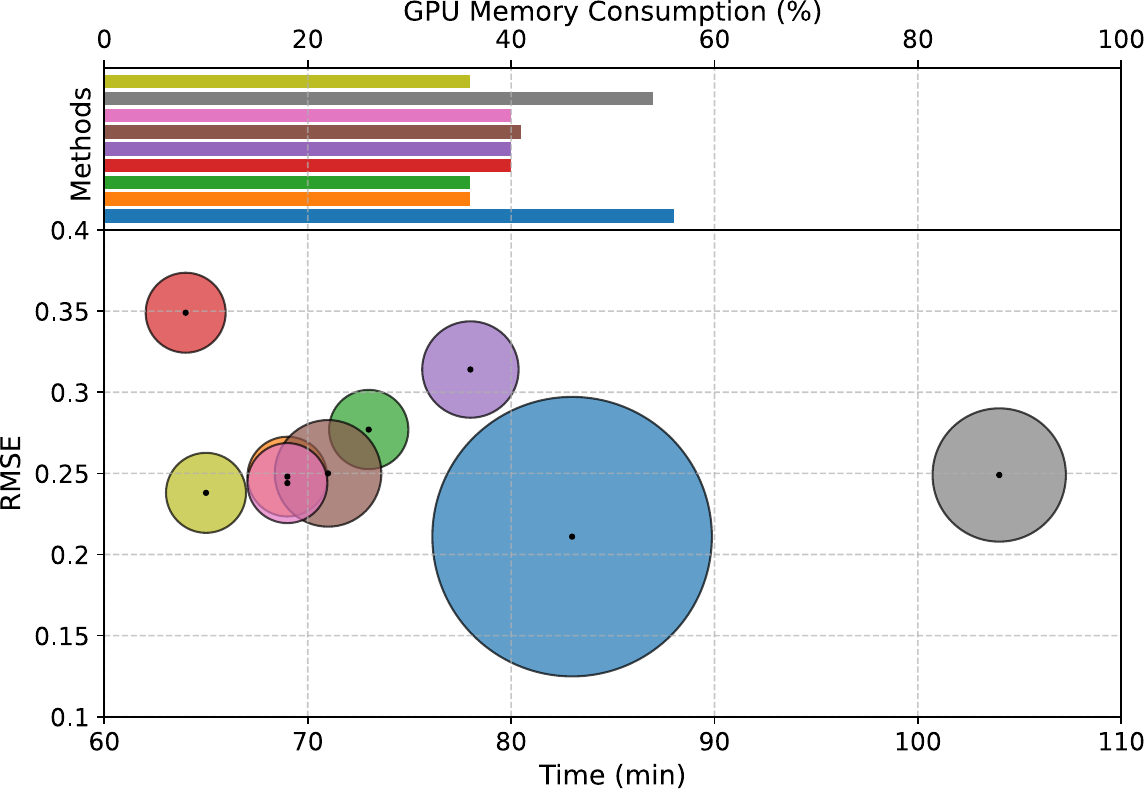}
         \caption{KITTI $\rightarrow$ NYU.}
         \label{fig:kitti_to_nyu}
     \end{subfigure}
     \begin{subfigure}[h]{0.49\textwidth}
         \centering
         \resizebox{\columnwidth}{!}{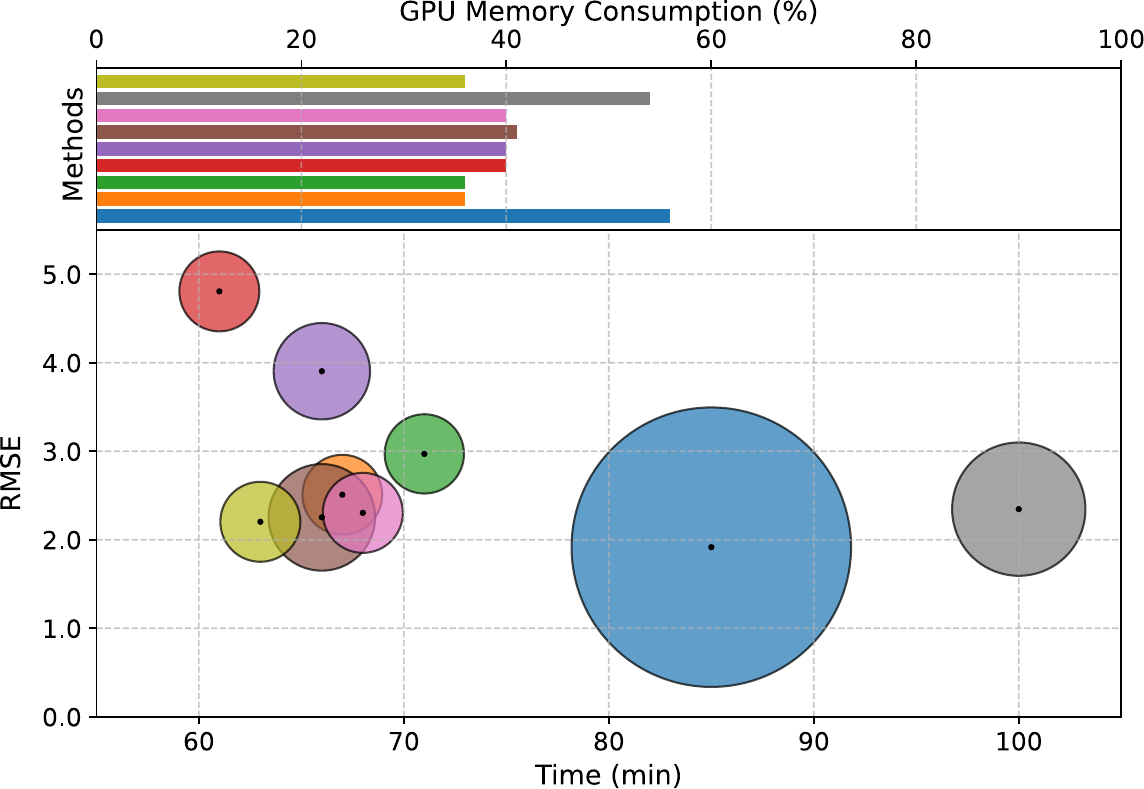}
         \caption{NYU $\rightarrow$ KITTI.}
         \label{fig:nyu_to_kitti}
     \end{subfigure}
     \begin{subfigure}[h]{0.49\textwidth}
         \centering
         \resizebox{\columnwidth}{!}{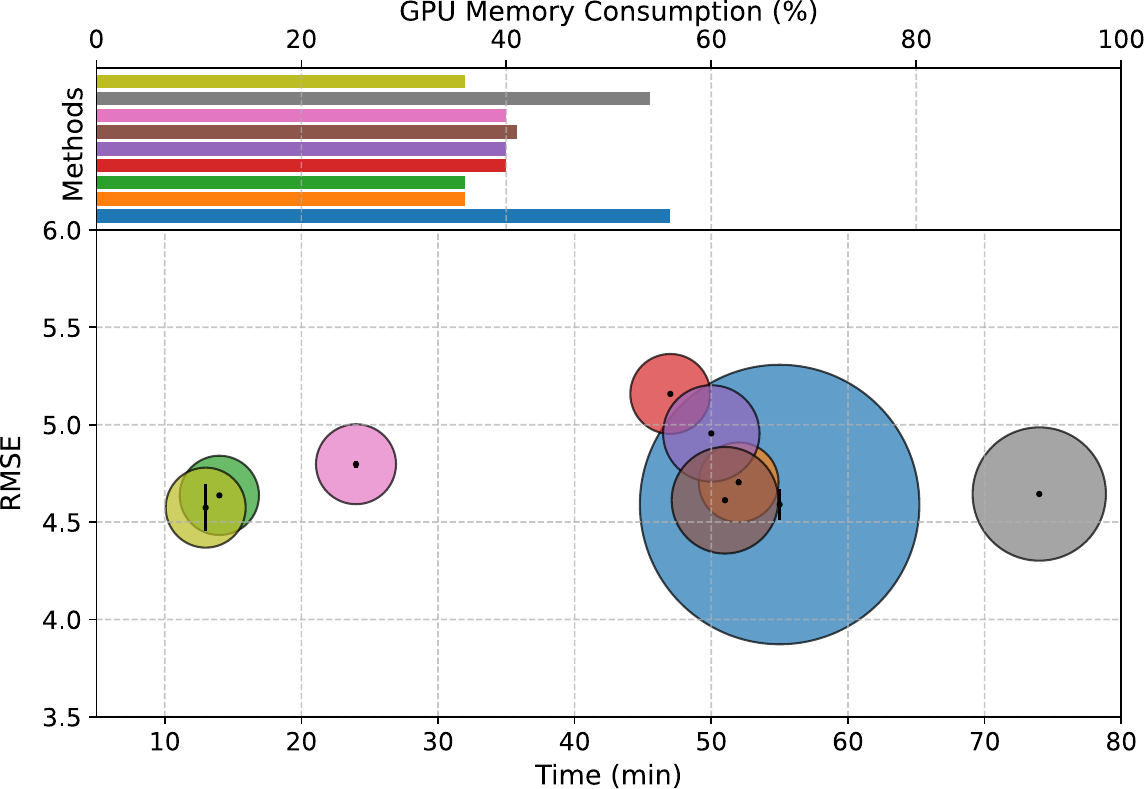}
         \caption{KITTI $\rightarrow$ DIODE Outdoor.}
         \label{fig:kitti_to_diodeout}
     \end{subfigure}
     \begin{subfigure}[h]{0.49\textwidth}
         \centering
         \resizebox{\columnwidth}{!}{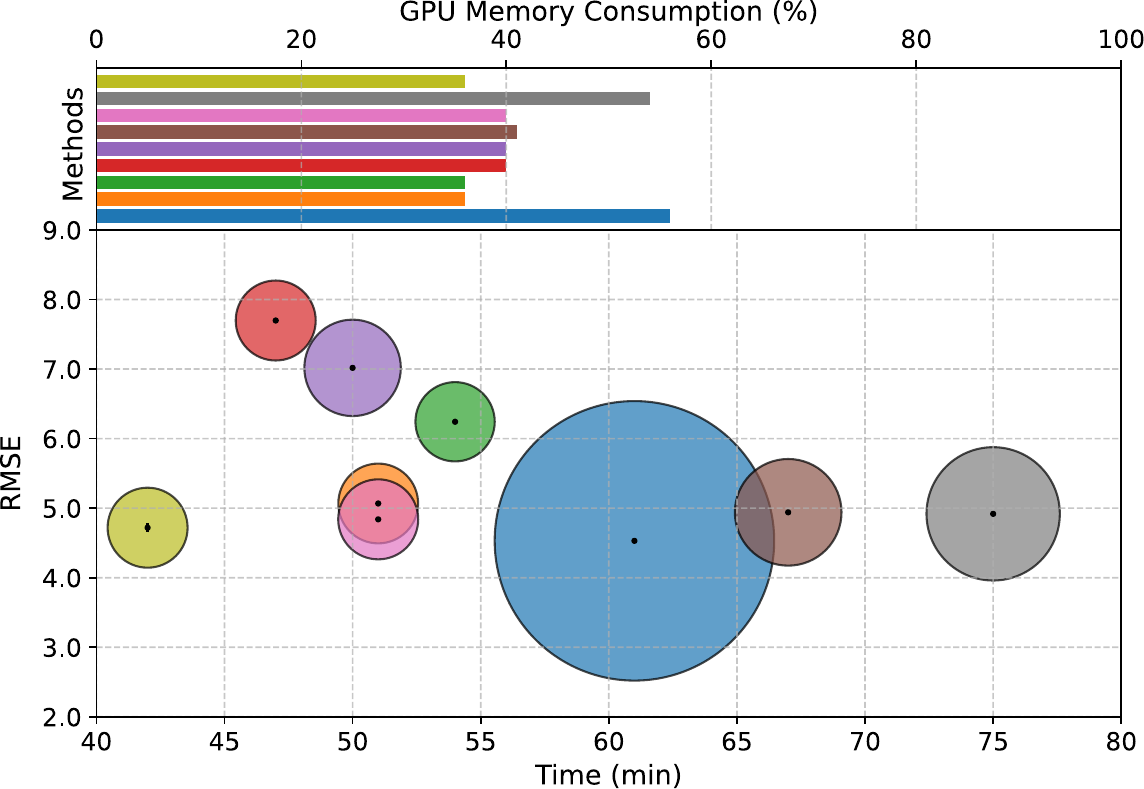}
         \caption{NYU $\rightarrow$ DIODE Outdoor.}
         \label{fig:nyu_to_diodeout}
     \end{subfigure}
    \caption{RMSE ($\downarrow$) of pre-trained DepthAnything model fine-tuned using PEFT methods. Bubble size is proportional to the number of parameters, except for "All", which is capped due to space limitations. Bottom left (and smallest bubble) is better.}
    \label{fig:mde}
\end{figure*}

\subsection{Monocular Depth Estimation}
As first experiment, we investigate the Monocular Depth Estimation (MDE) fine-tuning task. Here, we start from a pre-trained metric depth model and fine-tune it to a new dataset without introducing new untrained modules like adapters. We use the model and training settings of Depth Anything \cite{depthanything}, which are based on DINOv2 \cite{dinov2}, MiDaS \cite{midas,midas3}, DPT \cite{dpt} and ZoeDepth \cite{zoedepth}. We adopt widely used MDE benchmark datasets: NYU (indoor) \cite{nyu}, KITTI (driving) \cite{kitti} and DIODE (outdoor) \cite{diode}. We test the following fine-tuning cases: from NYU to KITTI, from KITTI to NYU, from NYU to DIODE Outdoor, from KITTI to DIODE Outdoor.
\Cref{tab:nyu_to_kitti} shows the Root Mean Squared Error (RMSE) and Absolute Relative Error (AbsRel) for the first two cases, respectively. Here, \ILoRA outperforms all the competitors and is the closest to fine-tuning the entire model. Additionally, \Cref{fig:kitti_to_nyu} and \Cref{fig:nyu_to_kitti} show the training time, number of parameters and GPU memory needed to achieve the minimum Root Mean Square Error (RMSE) for each method in the first two cases. 
From the figure, it emerges that not only is our method~\ilora{} the closest to full fine-tuning~\all{}, but it is also (1) faster than all competitors but BitFit~\bitfit, (2) it is the method with the lowest memory consumption together with MoRA~\mora, (3) it is the method fine-tuning the least parameters together with BitFit, MoRA and VeRA~\vera.

\Cref{fig:kitti_to_diodeout} and \Cref{fig:nyu_to_diodeout} show results for the fine-tuning cases from KITTI to DIODE Outdoor and from NYU to DIODE Outdoor, respectively. The figures confirm that our method~\ilora{} outperforms the competitors in terms of efficacy and computational efficiency, in terms of memory usage it is the best on par with MoRA~\mora{}, and in terms of parameters it is again the best on par with BitFit~\bitfit, MoRA and VeRA~\vera.

\subsection{Mathematical Reasoning}
\label{sec:math_reasoning}
We study the Mathematical Reasoning task, fine-tuning 7 billion and 13 billion parameters pre-trained LLaMA2 \cite{llama2} models to Meta-Math \cite{metamath}. We use the fine-tuning settings of MoRA \cite{jiang2024mora}, where for the 7 billion model we use 16 GPUs with 4 images each and a total batch size of 128 with gradient accumulation, and for the 13 billion model we use 64 GPUs with 2 images each. In addition, we split the dataset into 80\% for training and 20\% for validation.

\begin{figure}[tb]
    \resizebox{\columnwidth}{!}{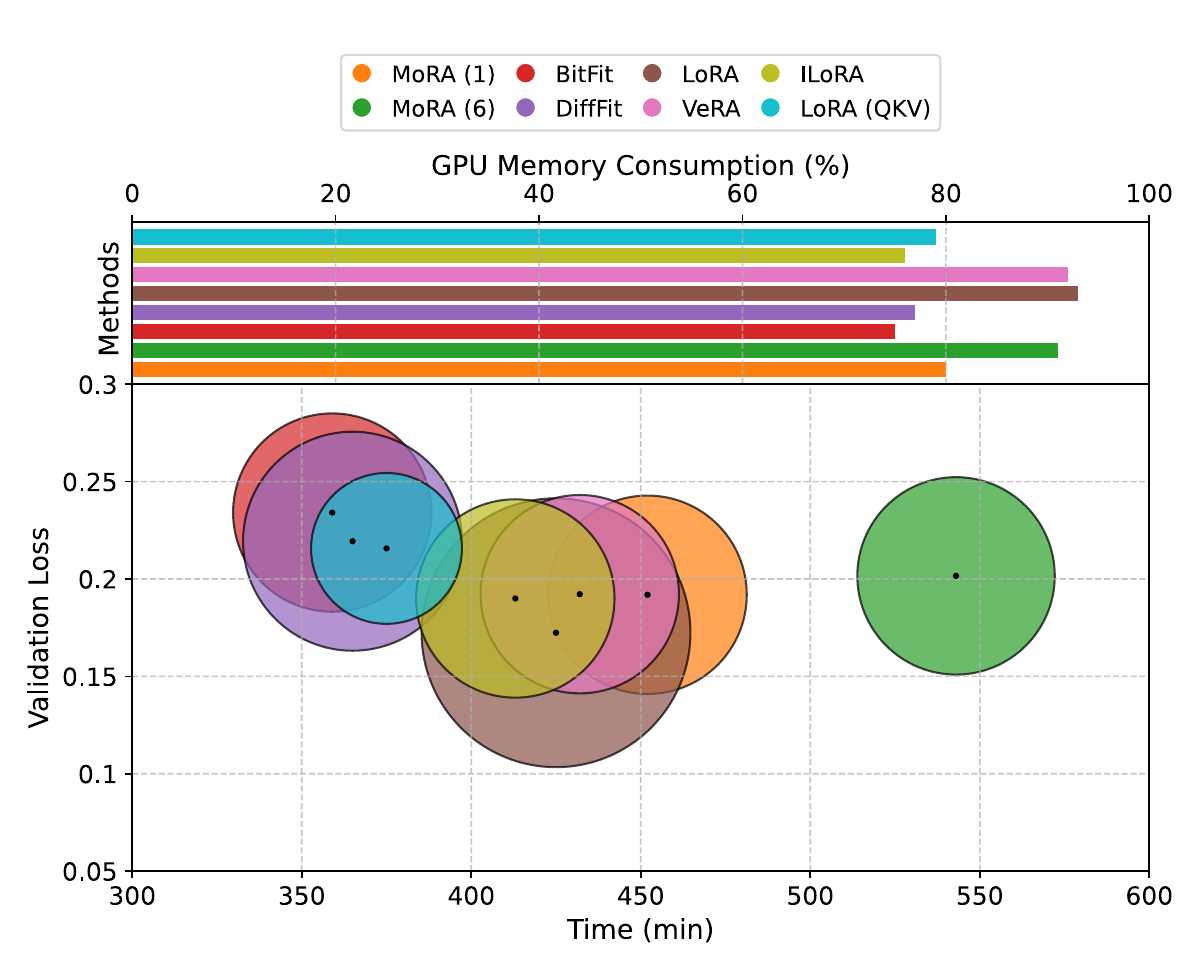}
    \caption{Validation loss of pre-trained LLaMA-2 7b fine-tuned to Meta-Math. Bubble size is proportional to the number of parameters. Bottom left (and smallest bubble) is better.} 
    \label{fig:llama2_7b}
\end{figure}

\Cref{fig:llama2_7b} shows that \ILoRA~\ilora{} outperforms all methods, except LoRA~\lora, in terms of validation loss, while using competitive computational time, parameter and memory budgets. LoRA achieves a slightly lower validation loss, while using more than $90\%$ of the GPU memory compared to less than $80\%$ of \ILoRA. On the other hand, when limiting the memory budget of LoRA by applying it only to the attention layers, here named LoRA (QKV)~\loraqkv, \ILoRA results in a better performance while still using slightly less GPU memory, showing that \ILoRA is more memory efficient than LoRA. Moreover, our method is only slower than BitFit~\bitfit, DiffFit~\difffit{} and LoRA (QKV), while being better in terms of quality. In terms of trainable parameters, \ILoRA is again comparable to BitFit and MoRA~\mora{}, and only second to LoRA (QKV), which is applied to fewer layers. Lastly, in terms of memory, \ILoRA is the second most efficient method, almost on par with the first one, BitFit.


\begin{figure}[tb]
    \resizebox{\columnwidth}{!}{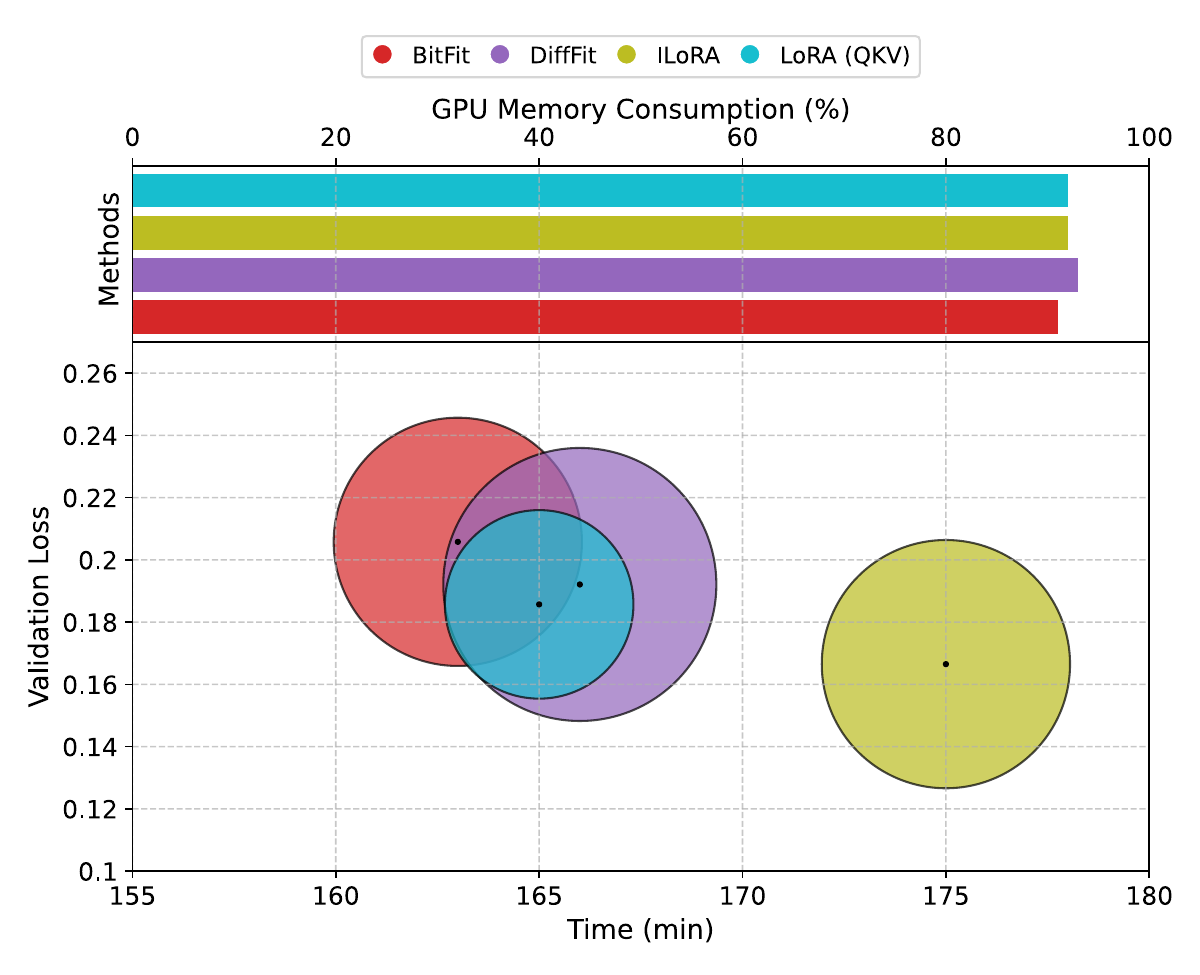}
    \caption{Validation loss of pre-trained LLaMA-2 13b fine-tuned to Meta-Math. Missing competitors cannot be fine-tuned with the given memory budget. Bubble size is proportional to the number of parameters. Bottom left (and smallest bubble) is better.} 
    \label{fig:llama2_13b}
\end{figure}

Given that \ILoRA is more memory efficient than LoRA and LoRA (QKV), \ILoRA can be applied to more layers and modules than LoRA in bigger models, allowing a more fine-grained fine-tuning than LoRA, which needs to be limited to specific layers. \Cref{fig:llama2_13b} exemplifies this \ILoRA benefit for fine-tuning a \mbox{LLaMA-2} 13b. LoRA cannot be applied to all linear layers as it would require too much GPU memory, therefore it has to be limited to specific layers (Q, K, V in this case). \ILoRA can be applied to all linear layers with a similar memory budget, while enabling a better performance. In fact, here, \ILoRA outperforms all applicable competitors while consuming a comparable memory budget and only being slightly slower. We report visual comparison of \ILoRA and LoRA QKV in Appendix~\ref{app:llama_visual}.

\subsection{Image Generation}
\begin{figure}[tb]
    \centering
    \resizebox{\columnwidth}{!}{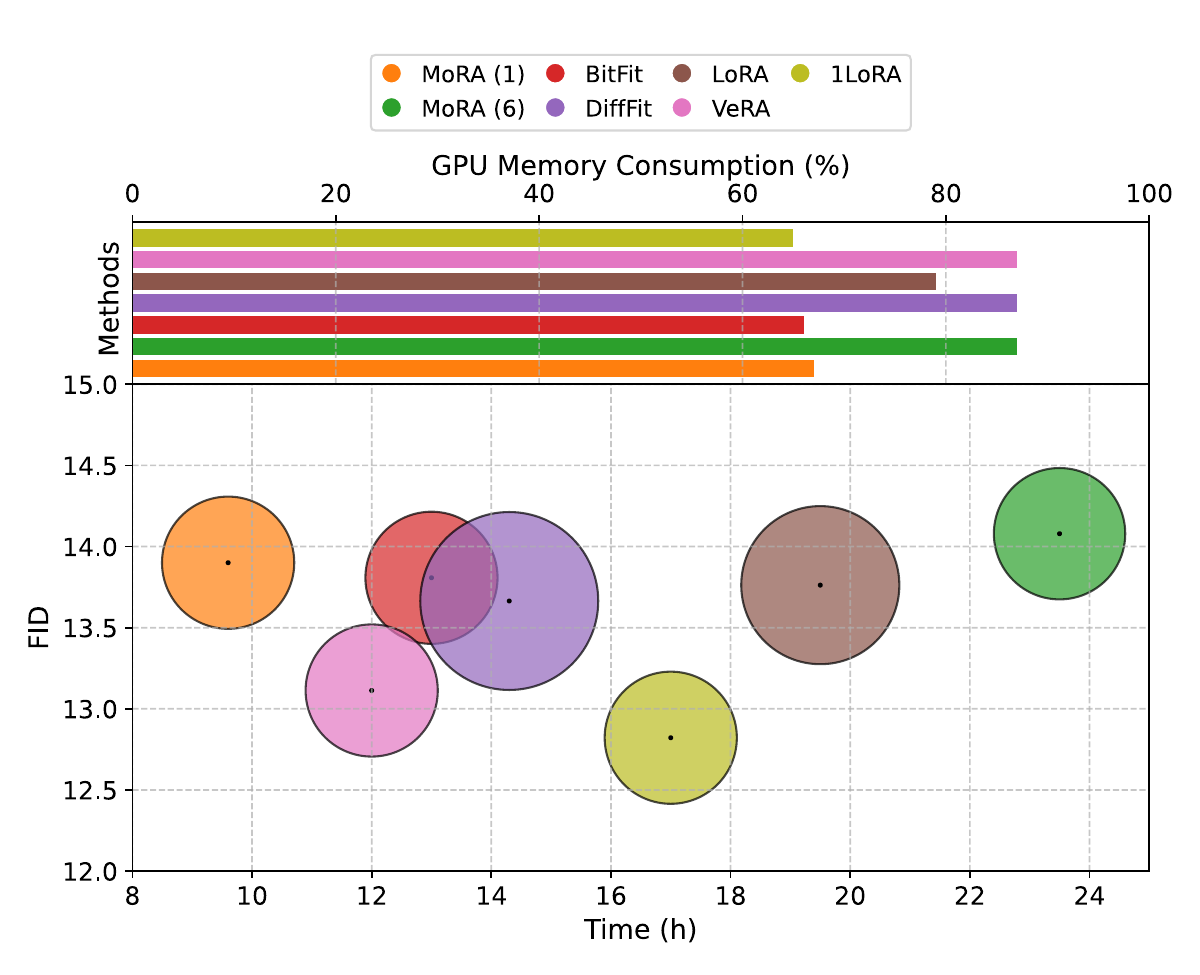}
    \caption{FID of pretrained DiT fine-tuned to Food-101. Note that DoRA is missing as it cannot be fine-tuned with the given memory budget. Bubble size is proportional to the number of parameters. Bottom left (and smallest bubble) is better.} 
    \label{fig:diffusion}
\end{figure}

We investigate the use of PEFT methods for diffusion model fine-tuning for image-generation. We fine-tune an ImageNet-21k \cite{ridnik2021imagenet21k} pre-trained DiT-XL-2-256x256 \cite{dit} to Food-101 \cite{food101} using the fine-tuning settings of DiffFit~\difffit{} \cite{xie2023difffit}. We run all experiments with a 24h time budget and we report results of a single run per method, given the high computational requirements. \Cref{fig:diffusion} shows that \ILoRA~\ilora{} outperforms all competitors in terms of FID, while being the most memory efficient method. Notice that \ILoRA also converges faster than LoRA~\lora{} and MoRA (6)~\morasix. In addition, as for previous experiments, \ILoRA has the lowest number of parameters together with BitFit~\bitfit, MoRA~\mora{} and VeRA~\vera. Lastly, DoRA~\dora{} cannot be fine-tuned with the same fine-tuning settings as it goes out of memory.

\subsection{Image Classification}
We compare \ILoRA and competitor methods on Image Classification. We use a ViT-Base \cite{vit} model pre-trained on ImageNet-21k \cite{ridnik2021imagenet21k} and we fine-tune it to CIFAR10 and CIFAR100 \cite{cifar}, respectively, using AdamW \cite{adamw} with learning rate $0.00002$, weight decay $0.01$, batch size 10, random resized crop to 224 and random horizontal flip. 

\begin{figure}[tb]
    \resizebox{\columnwidth}{!}{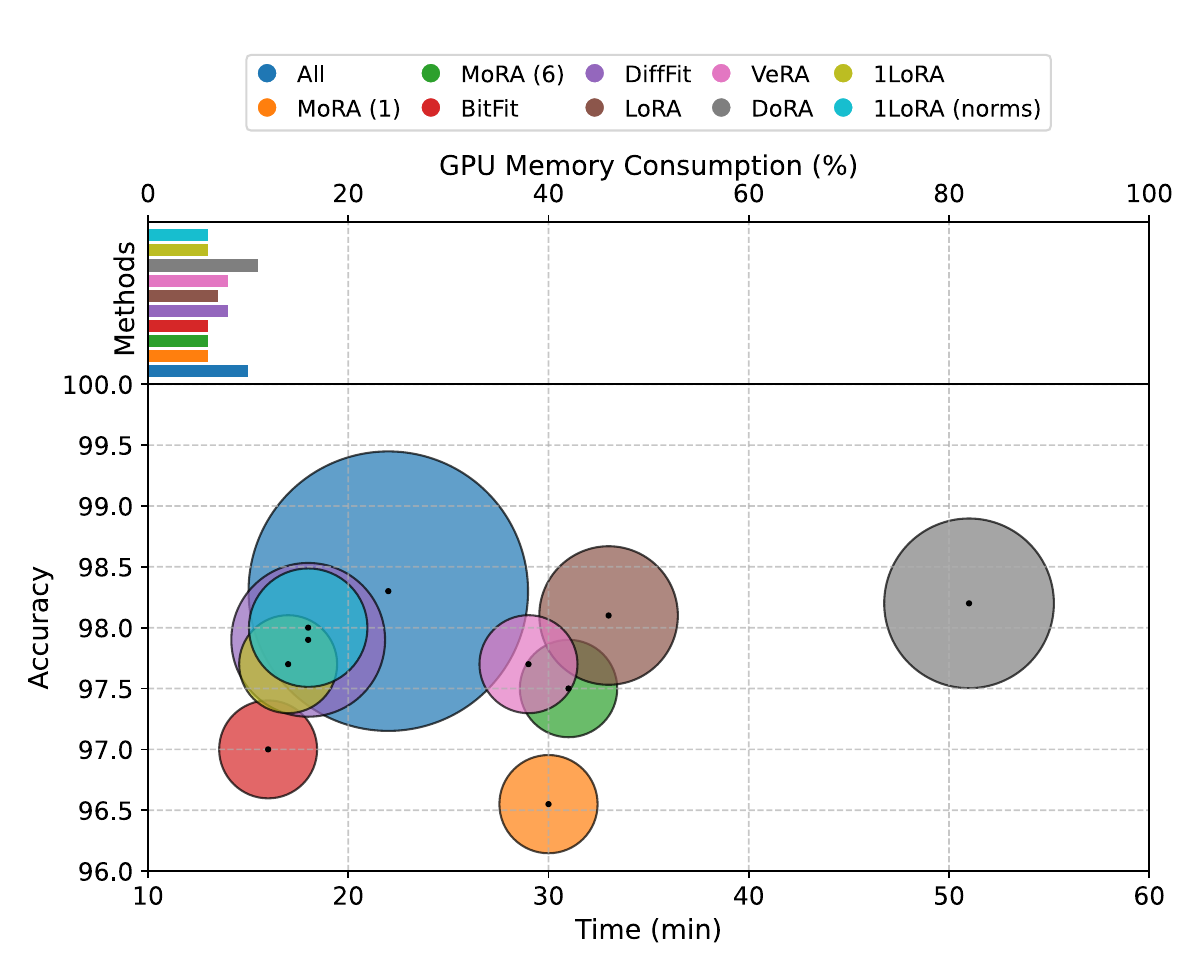}
    \caption{Accuracy ($\uparrow$) of ViT-Base model pre-trained on ImageNet-21k and fine-tuned to CIFAR10. Bubble size is proportional to the number of parameters, except for "All", which is capped. Top left (and smallest bubble) is better.} 
    \label{fig:imagenet_to_cifar10}
\end{figure}

\begin{figure}[tb]
    \resizebox{\columnwidth}{!}{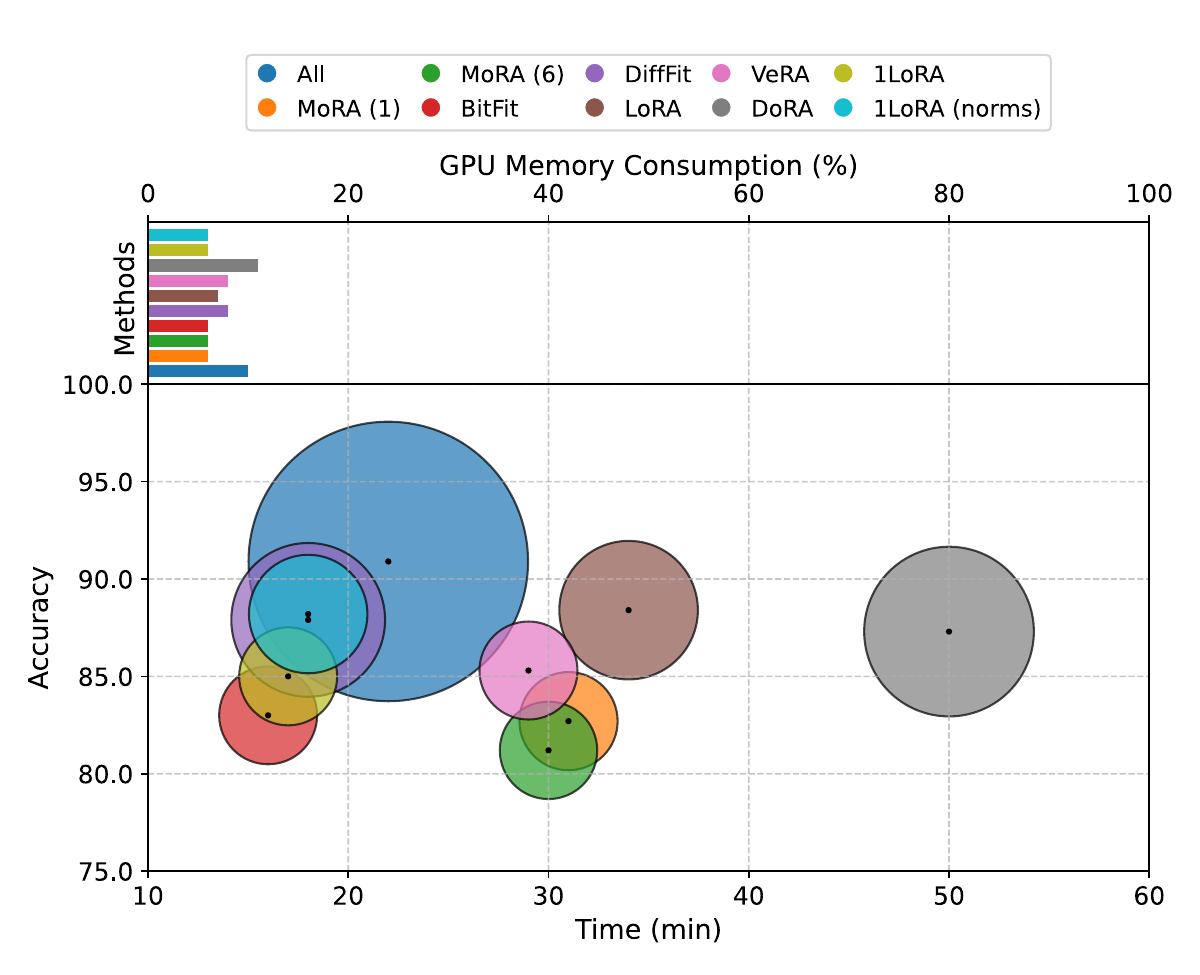}
    \caption{Accuracy ($\uparrow$) of ViT-Base model pre-trained on ImageNet-21k and fine-tuned to CIFAR100. Bubble size is proportional to the number of parameters, except for "All", which is capped. Top left (and smallest bubble) is better.} 
    \label{fig:imagenet_to_cifar100}
\end{figure}

\Cref{fig:imagenet_to_cifar10} and \Cref{fig:imagenet_to_cifar100} show the results for CIFAR10 and CIFAR100, respectively. From the figures, it emerges that \ILoRA~\ilora{} is the second fastest method after BitFit~\bitfit, while being slightly better in terms of accuracy. LoRA~\lora{} and DoRA~\dora{} are the methods achieving the closest accuracy to fine-tuning the entire model, however they are also the slowest methods, even slower than full fine-tuning. In this experiment, DiffFit~\difffit{} is slightly better than \ILoRA~\ilora{}, however it is also slightly slower and it uses more parameters and memory. It is worth noting that \ILoRA is complementary to BitFit~\bitfit and DiffFit~\difffit{}. They can be combined in many ways, as we study in \Cref{sec:complementary}. Here, the figures additionally report the most fruitful combination, \ILoRA (norms)~\iloranormsFigClassification, \ie\ \ILoRA and unfrozen normalization layers. This combination is slightly better, being more parameter and memory efficient than DiffFit. In \Cref{sec:complementary}, we study further combinations. Lastly, in order to show that fine-tuning the backbone leads to significant advantages, we report accuracies for the fine-tuning of the classification head only, on CIFAR10 (90\%) and CIFAR100 (66\%).

\section{Ablations}
\subsection{Complementary methods}
\label{sec:complementary}
We investigate all possible combinations of BitFit, DiffFit and \ILoRA, including individual components (the biases $\beta$, the scaling factors $\gamma$, and the norms, see \Cref{tab:baselines}), in order to understand which modules impact fine-tuning most. We run experiments for Image Classification, fine-tuning a ViT-Base from ImageNet-21k \cite{ridnik2021imagenet21k} to CIFAR10. We adopt the training configurations described in previous sections. 



\begin{figure}[t]
    \resizebox{\columnwidth}{!}{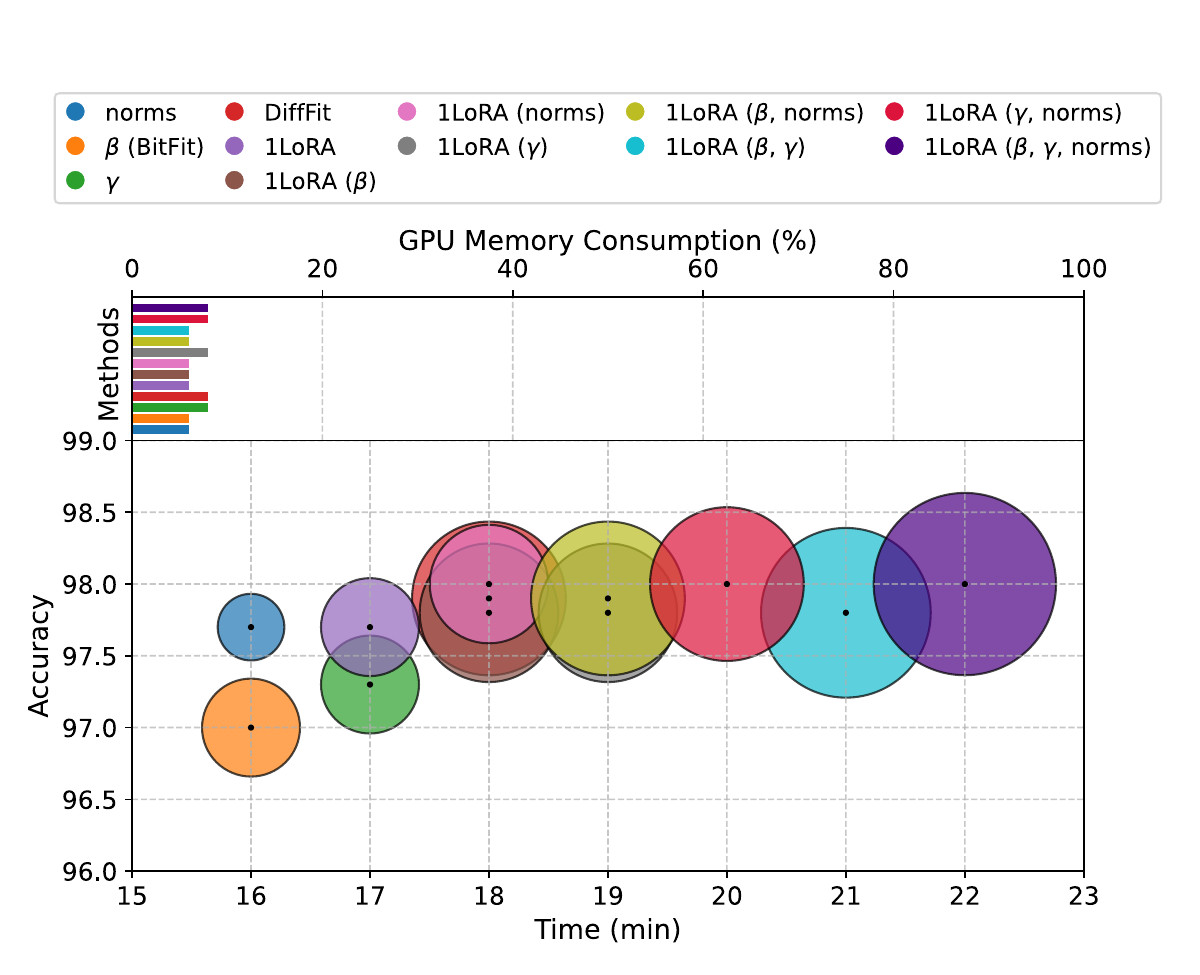}
    \caption{Accuracy ($\uparrow$) of ViT-Base model pre-trained on ImageNet-21k and fine-tuned to CIFAR10, using all combinations of \ILoRA, BitFit and DiffFit, including individual components (biases $\beta$, scaling factors $\gamma$ and normalization layers, $norms$). Bubble size is proportional to the number of parameters. Top left (and smallest bubble) is better.}
    \label{fig:ablation_complementary_cls}
\end{figure}

\Cref{fig:ablation_complementary_cls} reports results for the Image Classification task on CIFAR10, where there is a clear advantage in preferring fine-tuning of normalization layers~\norms{} over biases~\bias{} or scaling factors~\scaling. Combining these modules with \ILoRA leads to the best results, comparable to DiffFit~\difffit, but with fewer parameters, less memory consumption and smaller computational cost. Other combinations do not seem to bring additional performance gains. We report additional results for this ablation on the MDE task in Appendix~\ref{app:complementary}.

\subsection{PCA analysis}
\label{sec:pca_analysis}
\begin{figure}[t]
    \centering
    \includegraphics[width=0.8\columnwidth]{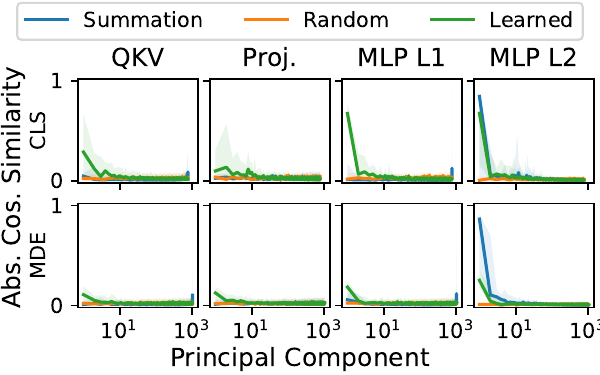}
    \caption{Alignment between PCs (sorted by singular values) of a full weight update and our candidate compression vectors.}
    \label{fig:pca}
\end{figure}

We performed PCA on the weight updates for our Monocular Depth Estimation and Classification experiments. 
We compare principal components (PCs) of a full-rank update against 
a random, summation (\ILoRA) and learned (LoRA, $r=1$) compression vector and present results in \Cref{fig:pca}.
For MDE, alignment of all compression vectors in terms of absolute cosine similarity to the PCs is mostly low, 
except for the second MLP layers.
For classification, a similar pattern is visible except that the learned compression achieves notably higher similarity,
especially also for the first MLP layer.
Interestingly, the second MLP layer of a ViT block is the only layer in the architecture, where features went through a GELU activation function right before-hand.
Effectively, the GELU activation concentrates the incoming features towards the positive ``quadrant'', thus summation (\ie~the $\mathds{1}$ vector) is unlikely to be orthogonal to those incoming features, and orthogonality may be an issue to performance as it suppresses the incoming feature.

As to why \ILoRA yields good performance, we argue that summation seems to be a good a priori guess for the PCs of linear layers
that are fed with GELU or ReLU activations. For the other layers, a fixed summation compression performs similarly to a random one. In Appendix~\ref{app:sum_vs_rand} we compare results for \ILoRA with fixed summation and random compression, showing that the former (\ie \ILoRA) is at least as good as a random compression, while being faster.
On the other hand, a learned compression only leads to slightly better alignment with the PCs (\cf~learned compression in \Cref{fig:pca}), while requiring double the amount of parameters and more memory.

In conclusion, given the similarity in compression quality, using 
\ILoRA
is preferable for memory and computational efficiencies.

\section{Conclusion}
PEFT methods have transformed the approach to fine-tuning large models for downstream tasks by enabling the adjustment of significantly fewer parameters than those in the original model matrices. 
In this work, we study the "very low rank regime", where we fine-tune the lowest amount of parameters per layer for each considered PEFT method. Inspired by LoRA, VeRA, and the recent MoRA, we propose \ILoRA, a compute, parameter and memory efficient fine-tuning method which uses the feature sum as compression, and a trainable vector as decompression. Differently from state-of-the-art PEFT methods, \ILoRA uses only $d$ parameters per layer, reducing the memory footprint and the computational cost. 
%

Moreover, by using the feature sum, instead of random projection, as input for the trainable decompression, \ILoRA makes one step forward in interpretable fine-tuning, suggesting that learning to shift the pre-trained parameters by scaled feature sum coming from samples drawn from a new dataset is enough to learn the new task.

We extensively evaluate our method against state-of-the-art PEFT methods on multiple fine-tuning tasks and models, namely Monocular Depth Estimation with DepthAnything, Image Classification with ViT-Base, Mathematical Reasoning with LLaMA2 7b and 13b, and Image Generation with DiT-XL-2-256x256. We observe that our method not only outperforms state-of-the-art methods, but is also more parameter, memory and computationally efficient. 

By reducing the number of trainable parameters per layer to $d$, \ILoRA enables the fine-tuning of big models (\eg\ LLMs with billions of parameters), where other methods cannot be applied due to their larger memory needs.
Additionally, we show that our method can be applied to more layers on bigger models, outperforming other competitors while using similar memory budget.

With this paper, we demonstrate the benefit of the very low-rank regime, where the lowest amount of parameters is fine-tuned, while maximizing the performance. 

\paragraph{Acknowledgements} 
A.Q. and R.D.P. were funded by the Helmholtz School for Data Science in Life, Earth, and Energy (HDS-LEE). The authors gratefully acknowledge the Gauss Centre for Supercomputing e.V. (\url{www.gauss-centre.eu}) for funding this project by providing computing time through the John von Neumann Institute for Computing (NIC) on the GCS Supercomputer JUWELS \cite{kesselheim2021juwels} at J\"ulich Supercomputing Centre (JSC).


{
    \small
    \bibliographystyle{ieeenat_fullname}
    \bibliography{main}
}

\appendix
\setcounter{table}{0}
\renewcommand{\thetable}{S\arabic{table}}
\clearpage
\twocolumn[
\centering
\textbf{ \Large Supplemental Material\\ \ILoRA: Summation Compression for Very Low-Rank Adaptation\vspace*{\baselineskip}\\}]

\section{Algorithm}
As stated in the main paper, \ILoRA is straightforward to implement. We illustrate the python implementation in Listing \ref{py_ilora}. In our work, we applied the \ILoRA module to all linear layers in the model, excluding the classification layers, when present.

\begin{lstlisting}[language=Python, caption=Implementation of \ILoRA in Python, label=py_ilora]
import torch    
import torch.nn as nn

class ILoRA(nn.Module):
    def __init__(self, linear):
        super().__init__()
        self.linear = linear
        self.b = nn.Parameter(torch.zeros(self.linear.out_features))
   
    def forward(self, x):
        ilora = self.b * x.sum(-1, keepdim=True)
        return self.linear(x) + ilora
\end{lstlisting}

\section{Ablations}
\subsection{Complementary methods}
\label{app:complementary}
We investigate all possible combinations of BitFit, DiffFit and \ILoRA, including individual components (the biases $\beta$, the scaling factors $\gamma$, and the norms, see \Cref{tab:baselines}), in order to understand which modules impact fine-tuning most. We report experiments for MDE, fine-tuning DepthAnything from KITTI to NYU. 

\begin{figure}[t]
    \resizebox{\columnwidth}{!}{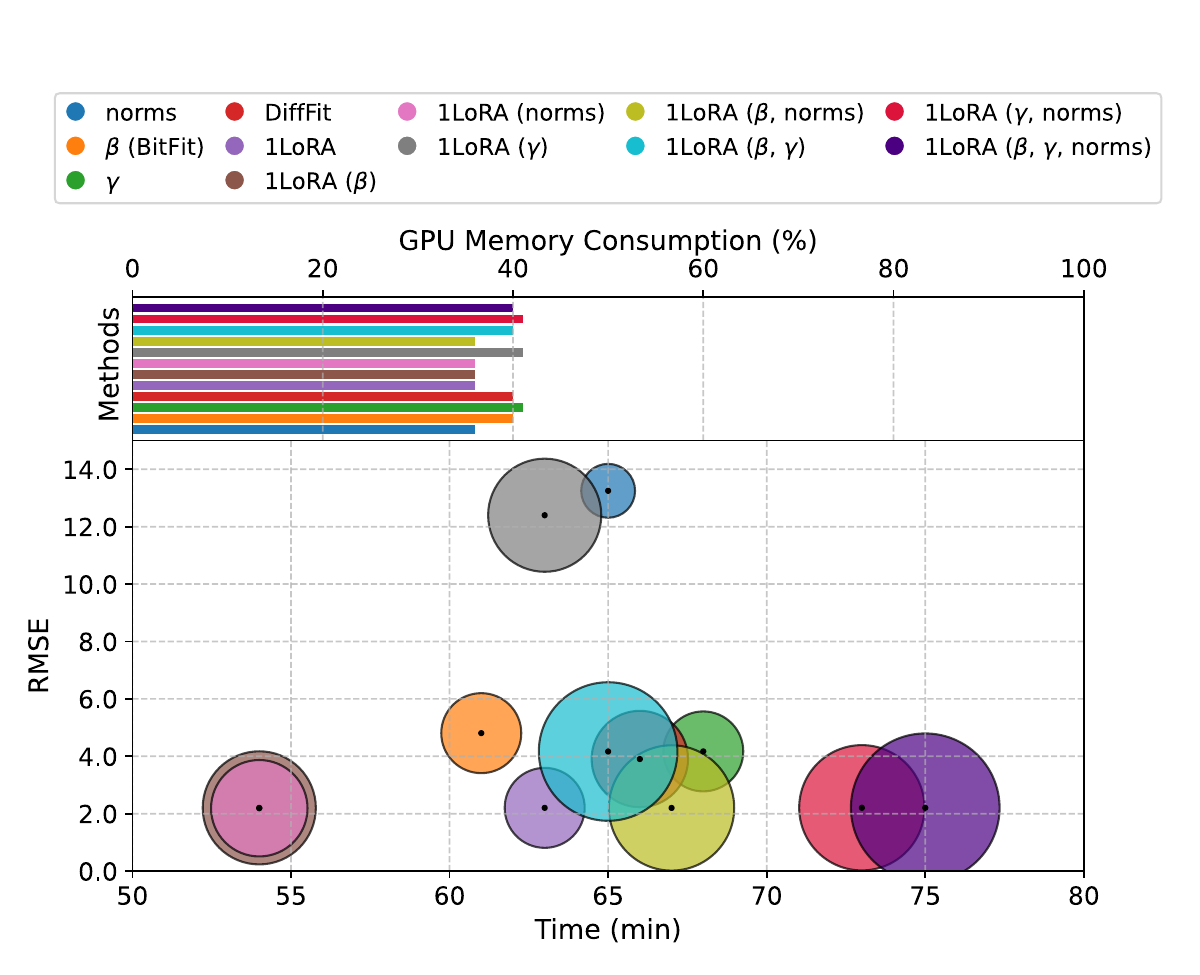}
    \caption{RMSE ($\downarrow$) of DepthAnything model pre-trained on KITTI and fine-tuned to NYU, using all combinations of \ILoRA, BitFit and DiffFit, including individual components (biases $\beta$, scaling factors $\gamma$ and normalization layers $norms$). Bubble size is proportional to the number of parameters. Bottom left (and smallest bubble) is better.}
    \label{fig:ablation_complementary_mde}
\end{figure}

Interestingly, \Cref{fig:ablation_complementary_mde} shows that, for the MDE case, only fine-tuning the normalization layers (norms, \norms) leads to suboptimal results, whereas combining it with \ILoRA~\iloranorms{} allows it to achieve similar results as \ILoRA~\ilorac, but faster. Fine-tuning the biases alone (i.e.\ BitFit, \bias) achieves below state-of-the-art performance, but combining it with \ILoRA~\ilorabias{} has the same effect as combining norms with \ILoRA~\iloranorms. Other combinations seem to lead to no additional benefit, and only slow down the fine-tuning process. In general, any combination also increases the amount of trainable parameters, and \ILoRA already achieves the best results while using the least number of trainable parameters in this experiment.

\subsection{Summation vs random compression}
\label{app:sum_vs_rand}
Here, we report ablation on the MDE (Table~\ref{tab:ilora_rand_mde}) taskcomparing \ILoRA with fixed summation and random compression. Results show that summation is overall slightly better in terms of performance or required time, suggesting that using a fixed summation compression is at least as good as using random compression.

\begin{table}[tbh!]
    \centering
    \caption{(MDE) \ILoRA: summation vs random compression.}
    \small


\setlength{\tabcolsep}{1pt}
\begin{tabular}{L{3.cm} C{1.6cm} C{1.2cm} C{2.2cm}}
\toprule
  dataset & compression & time [min] ($\downarrow$) & RMSE ($\downarrow$) \\
\midrule
   \multirow{2}{*}{NYU $\rightarrow$ KITTI} & summation & $\bm{63}$ & $\bm{2.203 \pm 0.0078}$  \\
    & random & 65 & 2.239 $\pm$ 0.0089  \\
   \midrule
   \multirow{2}{*}{KITTI $\rightarrow$ NYU} & summation & $\bm{65}$ & $\bm{0.238 \pm 0.0005}$  \\
    & random & 67 & 0.24 $\pm$ 0.0002  \\
   \midrule
   \multirow{2}{*}{KITTI $\rightarrow$ DIODE Out} & summation & $\bm{13}$ & $\bm{4.574 \pm 0.1217}$  \\
    & random & 22 & 4.573 $\pm$ 0.0582  \\
   \midrule
   \multirow{2}{*}{NYU $\rightarrow$ DIODE Out} & summation & 42 & $\bm{4.719 \pm 0.06}$  \\
    & random & 42 & 4.725 $\pm$ 0.06 \\
\bottomrule
\end{tabular}
    \label{tab:ilora_rand_mde}
\end{table}


\subsection{Analysis of edge cases}
\begin{figure}[t]
    \centering
    \includegraphics[width=0.8\columnwidth]{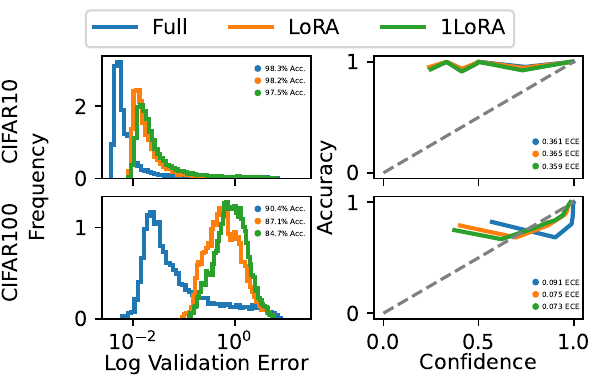}
    \caption{Error distribution and calibration for classification examples.}
    \label{fig:errdist}
\end{figure}

To explore edge cases, we analyzed the error distribution
and calibration of full fine-tuning, LoRA ${r=1}$ and \ILoRA for our classification experiments (Fig.~\ref{fig:errdist}).
Contrary to expectation, full fine-tuning seems to be more heavy-tailed in terms of error distribution than the very low-rank
PEFT methods.
As for the calibration, again the very low-rank PEFT methods seem to slightly improve above full fine-tuning, though
the improvement -- if any -- is very much negligible.

\section{Mathematical reasoning visualization}
\label{app:llama_visual}
We report visualization for LLaMA2 13b finetuned on 80\% MetaMath, comparing \ILoRA and LoRA QKV. We use random samples from our generated validation split (20\%) of MetaMath. We use black, olive and cyan boxes to highlight the input prompt (question and answer, where only the question is given as input and the answer serves as ground truth), the \ILoRA and LoRA QKV answers, respectively. Note that we omit ``Below is an instruction that describes a task. Write a response that appropriately completes the request." from Q \& A boxes due to space limitations.


\begin{tcolorbox}[question]

\#\#\# Instruction:

On Friday, a snowboard originally priced at $\$100$ was discounted $50\%$. On Monday, that sale price was reduced by $30\%$. In dollars, what is the price of the snowboard after the Monday reduction?

\#\#\# Response: 

The Friday sale price was $\$100 \times 0.50 = \$50$.
The Monday reduction is $30\%$ of $\$50$, so it is $\$50 \times 0.30 = \$15$.
Therefore, the price of the snowboard after the Monday reduction is $\$50 - \$15 = \boxed{\$35}$.The answer is: 35.
\end{tcolorbox}

\begin{tcolorbox}[ilora]
The Friday discount of $50\%$ reduces the price to $100 - 0.5(100) = 50$.
The Monday reduction of $30\%$ reduces the price further to $50 - 0.3(50) = \boxed{35}$.
The answer is: 35
\end{tcolorbox}

\begin{tcolorbox}[loraqkv]
The snowboard was originally priced at $\$100$.
The $50\%$ discount brings the price down to $\$100 \cdot 0.5 = \$50$.
The $30\%$ reduction brings the price down further to $\$50 \cdot 0.7 = \$35$.
Therefore, the price of the snowboard after the Monday reduction is $\boxed{\$35}$.The answer is: 35
\end{tcolorbox}

\begin{tcolorbox}[question]

\#\#\# Instruction:

George went to a movie theater to see a movie. He paid \$16 for the ticket and decided to buy some nachos. The nachos were half the price of the ticket. How much did George pay in total for the visit to the movie theater?

\#\#\# Response: 

George paid \$16 for the ticket.
The nachos were half the price of the ticket, so they cost \$16/2 = \$8.
In total, George paid \$16 + \$8 = \$24 for the visit to the movie theater.
\end{tcolorbox}

\begin{tcolorbox}[ilora]
The nachos were half the price of the ticket, so they cost \$16/2 = \$8.
Therefore, George paid a total of \$16 + \$8 = \$24 for the visit to the movie theater.

\#\#\#\# 24

The answer is: 24
\end{tcolorbox}

\begin{tcolorbox}[loraqkv]
The nachos were half the price of the ticket, so they cost \$16/2 = \$8
In total, George paid \$16 + \$8 = \$24 for the visit to the movie theater.

\#\#\#\# 24

The answer is: 24
\end{tcolorbox}

\begin{tcolorbox}[question]

\#\#\# Instruction:

If $ab + bc + cd + da = 30$ and $b + d = 5$, what is the value of $a + c$?

\#\#\# Response: 

We can rewrite the given equation as $a(b+d) + c(b+d) = 30$.
Using the second equation, we substitute $b+d = 5$ to get $5a + 5c = 30$.
Dividing both sides by 5, we have $a+c = \boxed{6}$.
The answer is: 6
\end{tcolorbox}

\begin{tcolorbox}[ilora]
We can rewrite the equation $ab + bc + cd + da = 30$ as $(a+c)(b+d) = 30$.
Substituting $b+d = 5$, we have $(a+c)(5) = 30$.
Dividing both sides by 5, we get $a+c = \boxed{6}$.
The answer is: 6
\end{tcolorbox}

\begin{tcolorbox}[loraqkv]
We can rewrite the given equation as $a(b+d) + c(b+d) = 30$.
Using the given information that $b+d=5$, we have $2ab + 2cd = 30$.
We can rewrite this equation as $ab + cd = 15$.
Since $ab + bc + cd + da = 30$, we can subtract $ab + cd$ from both sides to get $bc + da = 15$.
We can also rewrite this equation as $bc + da = 15$.
Since $b + d = 5$, we can substitute this into the equation to get $bc + da = 5$.
Now, we can solve for $a + c$ by subtracting $a$ and $c$ from both sides: $a + c = 5$.
Therefore, the value of $a + c$ is $\boxed{5}$.
The answer is: 5
\end{tcolorbox}

\begin{tcolorbox}[question]

\#\#\# Instruction:

In 5 years, Raven will be 4 times as old as Phoebe. If Phoebe is currently x years old, Raven is 55 years old. What is the value of unknown variable x?

\#\#\# Response: 

To solve this problem, we need to determine the value of x, which represents Phoebe's current age.
Let's break down the information given:
In 5 years, Raven will be 4 times as old as Phoebe.
Raven is currently 55 years old.
We can set up the equation as follows:
Raven's age in 5 years = 4 * Phoebe's age in 5 years
(Raven's current age + 5) = 4 * (Phoebe's current age + 5)
55 + 5 = 4 * (x + 5)
60 = 4 * (x + 5)
Let's simplify and solve for x:
60 = 4x + 20
To isolate x, we subtract 20 from both sides of the equation:
60 - 20 = 4x + 20 - 20
40 = 4x
To solve for x, we divide both sides of the equation by 4:
40 / 4 = 4x / 4
10 = x
The value of x is 10.

\#\#\#\# 10

The answer is: 10.
\end{tcolorbox}

\begin{tcolorbox}[ilora]
To solve this problem, we need to determine the value of x, which represents Phoebe's current age.
We know that in 5 years, Raven will be 4 times as old as Phoebe. Therefore, Raven's age in 5 years will be 4 times Phoebe's age in 5 years.
Let's break down the information given:
Raven's current age: 55 years
Phoebe's current age: x years
In 5 years, Raven's age: 55 + 5 = 60 years
In 5 years, Phoebe's age: x + 5 years
We can set up the equation as follows:
Raven's age in 5 years = 4 * Phoebe's age in 5 years
60 = 4 * (x + 5)
Let's simplify and solve for x:
60 = 4x + 20
To isolate x, we subtract 20 from both sides of the equation:
60 - 20 = 4x + 20 - 20
40 = 4x
To solve for x, we divide both sides of the equation by 4:
40 / 4 = 4x / 4
10 = x
The value of x is 10.

\#\#\#\# 10

The answer is: 10
\end{tcolorbox}

\begin{tcolorbox}[loraqkv]
To solve this problem, we need to determine the value of x, which represents Phoebe's current age.
We know that in 5 years, Raven will be 4 times as old as Phoebe. Therefore, Raven will be 4x years old in 5 years.
We also know that Raven is currently 55 years old.
We can set up the equation as follows:
Raven's age in 5 years = 4 * Phoebe's current age
55 = 4 * x
Let's simplify and solve for x:
55 = 4x
To isolate x, we divide both sides of the equation by 4:
55 / 4 = 4x / 4
14 = x
The value of x is 14.

\#\#\#\# 14

The answer is: 14
\end{tcolorbox}

\clearpage
\pagebreak

\onecolumn

\begin{minipage}{0.45\textwidth}    
\section{Detailed results}
We report tables (from \ref{tab:kitti_to_nyu_all_err} to \ref{tab:pt_to_cifar10_complementary_err}) illustrating the detailed results shown in the main paper as bubble and bar plots. For each column reporting a result, we highlight the best value in \textbf{bold}, the second best in \underline{underlined} and the third in \textit{italic}.
\end{minipage}

\begin{table*}[h!]
    \centering

\newrobustcmd{\rankfirst}{\fontseries{b}\selectfont}
\newrobustcmd{\rankthird}{\fontshape{sl}\selectfont}
\setlength{\tabcolsep}{5pt}
\begin{tabular}{L{1.cm} C{1.3cm} C{2.5cm} C{1.1cm} C{1.1cm}}
\toprule
  method & time [min] ($\downarrow$) & RMSE ($\downarrow$) & memory [\%] ($\downarrow$) & params ($\downarrow$) \\
\midrule
MoRA$_1$ &               \textit{69} &       \textit{0.248} $\pm$ \textit{0.00021} &                \textbf{36} &   \underline{220.06K} \\
MoRA$_6$ &                        73 &                         0.277 $\pm$ 0.00066 &                \textbf{36} &      \textbf{217.44K} \\
  BitFit &               \textbf{64} &                         0.349 $\pm$ 0.00073 &             \underline{40} &      \textit{221.18K} \\
 DiffFit &                        78 &                         0.314 $\pm$ 0.00078 &             \underline{40} &               321.54K \\
    LoRA &                        71 &                          0.25 $\pm$ 0.00087 &                \textit{41} &               393.22K \\
    VeRA &               \textit{69} & \underline{0.244} $\pm$ \underline{0.00059} &             \underline{40} &               221.28K \\
    DoRA &                       104 &                         0.249 $\pm$ 0.00088 &                         54 &               614.40K \\
   \ILoRA &            \underline{65} &        \textbf{0.238} $\pm$ \textbf{0.0005} &                \textbf{36} &      \textit{221.18K} \\
\midrule
     All &                        83 &                         0.211 $\pm$ 0.00064 &                         56 &               335.32M \\
\bottomrule
\end{tabular}
    \caption{Table for \Cref{fig:kitti_to_nyu}}
    \label{tab:kitti_to_nyu_all_err}
\end{table*}

\begin{table*}[tbh]
    \centering

\setlength{\tabcolsep}{5pt}
\begin{tabular}{L{1.cm} C{1.3cm} C{2.5cm} C{1.1cm} C{1.1cm}}
\toprule
  method & time [min] ($\downarrow$) & RMSE ($\downarrow$) & memory [\%] ($\downarrow$) & params ($\downarrow$) \\
\midrule
MoRA$_1$ &                        67 &                         2.509 $\pm$ 0.00455 &                \textbf{36} &   \underline{220.06K} \\
MoRA$_6$ &                        71 &                          2.97 $\pm$ 0.00328 &                \textbf{36} &      \textbf{217.44K} \\
  BitFit &               \textbf{61} &                         4.806 $\pm$ 0.00392 &             \underline{40} &      \textit{221.18K} \\
 DiffFit &               \textit{66} &                         3.904 $\pm$ 0.00539 &             \underline{40} &               321.54K \\
    LoRA &               \textit{66} & \underline{2.253} $\pm$ \underline{0.00315} &                \textit{41} &               393.22K \\
    VeRA &                        68 &       \textit{2.304} $\pm$ \textit{0.00524} &             \underline{40} &               221.28K \\
    DoRA &                       100 &                         2.346 $\pm$ 0.00275 &                         54 &               614.40K \\
   \ILoRA &            \underline{63} &        \textbf{2.203} $\pm$ \textbf{0.0078} &                \textbf{36} &      \textit{221.18K} \\
\midrule
     All &                        85 &                         1.916 $\pm$ 0.00303 &                         56 &               335.32M \\
\bottomrule
\end{tabular}

    \caption{Table for \Cref{fig:nyu_to_kitti}}
    \label{tab:nyu_to_kitti_all_err}
\end{table*}

\begin{table*}[tbh]
    \centering

\setlength{\tabcolsep}{5pt}
\begin{tabular}{L{1.cm} C{1.3cm} C{2.5cm} C{1.1cm} C{1.1cm}}
\toprule
  method & time [min] ($\downarrow$) & RMSE ($\downarrow$) & memory [\%] ($\downarrow$) & params ($\downarrow$) \\
\midrule
MoRA$_1$ &                        52 &                         4.704 $\pm$ 0.01546 &                \textbf{36} &   \underline{220.06K} \\
MoRA$_6$ &            \underline{14} &       \textit{4.637} $\pm$ \textit{0.00449} &                \textbf{36} &      \textbf{217.44K} \\
  BitFit &                        47 &                         5.158 $\pm$ 0.00868 &             \underline{40} &      \textit{221.18K} \\
 DiffFit &                        50 &                          4.955 $\pm$ 0.0064 &             \underline{40} &               321.54K \\
    LoRA &                        51 & \underline{4.612} $\pm$ \underline{0.00368} &                \textit{41} &               393.22K \\
    VeRA &               \textit{24} &                          4.797 $\pm$ 0.0181 &             \underline{40} &               221.28K \\
    DoRA &                        74 &                         4.644 $\pm$ 0.00405 &                         54 &               614.40K \\
   \ILoRA &               \textbf{13} &        \textbf{4.574} $\pm$ \textbf{0.1217} &                \textbf{36} &      \textit{221.18K} \\
\midrule
     All &                        55 &                           4.59 $\pm$ 0.0784 &                         56 &               335.32M \\
\bottomrule
\end{tabular}
    \caption{Table for \Cref{fig:kitti_to_diodeout}}
    \label{tab:kitti_to_diodeout_all_err}
\end{table*}

\begin{table*}[tbh]
    \centering

\setlength{\tabcolsep}{5pt}
\begin{tabular}{L{1.cm} C{1.3cm} C{2.5cm} C{1.1cm} C{1.1cm}}
\toprule
  method & time [min] ($\downarrow$) & RMSE ($\downarrow$) & memory [\%] ($\downarrow$) & params ($\downarrow$) \\
\midrule
MoRA$_1$ &                        51 &                         5.067 $\pm$ 0.02273 &                \textbf{36} &   \underline{220.06K} \\
MoRA$_6$ &                        54 &                          6.242 $\pm$ 0.0141 &                \textbf{36} &      \textbf{217.44K} \\
  BitFit &            \underline{47} &                         7.698 $\pm$ 0.00293 &             \underline{40} &      \textit{221.18K} \\
 DiffFit &               \textit{50} &                         7.017 $\pm$ 0.00998 &             \underline{40} &               321.54K \\
    LoRA &                        67 &                          4.94 $\pm$ 0.00776 &                \textit{41} &               393.22K \\
    VeRA &                        51 & \underline{4.839} $\pm$ \underline{0.01025} &             \underline{40} &               221.28K \\
    DoRA &                        75 &       \textit{4.918} $\pm$ \textit{0.00949} &                         54 &               614.40K \\
   \ILoRA &               \textbf{42} &         \textbf{4.719} $\pm$ \textbf{0.066} &                \textbf{36} &      \textit{221.18K} \\
\midrule
     All &                        61 &                           4.53 $\pm$ 0.0278 &                         56 &               335.32M \\
\bottomrule
\end{tabular}
    \caption{Table for \Cref{fig:nyu_to_diodeout}}
    \label{tab:nyu_to_diodeout_all_err}
\end{table*}

\begin{table*}[tbh]
    \centering

\setlength{\tabcolsep}{5pt}
\begin{tabular}{L{2.cm} C{1.3cm} C{2.5cm} C{1.1cm} C{1.1cm}}
\toprule
    method & time [min] ($\downarrow$) & Loss ($\downarrow$) & memory [\%] ($\downarrow$) & params ($\downarrow$) \\
\midrule
  MoRA$_1$ &                       187 &                         0.18 $\pm$ 0.000724 &                         80 &                 1.36M \\
  MoRA$_6$ &                       198 &                          0.19 $\pm$ 6.6e-05 &                         91 &     \underline{1.35M} \\
    BitFit &              \textbf{130} &                           0.22 $\pm$ 0.0047 &                \textbf{75} &        \textit{1.36M} \\
   DiffFit &           \underline{132} &                          0.21 $\pm$ 0.00411 &                \textit{77} &                 1.66M \\
      LoRA &                       154 &        \textbf{0.15} $\pm$ \textbf{0.00202} &                         93 &                 2.50M \\
      VeRA &                       159 &       \textit{0.176} $\pm$ \textit{0.00192} &                         92 &                 1.36M \\
     \ILoRA &                       152 & \underline{0.17} $\pm$ \underline{0.003768} &             \underline{76} &        \textit{1.36M} \\
LoRA (QKV) &              \textit{134} &                          0.2 $\pm$ 0.002273 &                         79 &      \textbf{786.43K} \\
\bottomrule
\end{tabular}

    \caption{Table for \Cref{fig:llama2_7b}}
    \label{tab:llama2_7b_to_metamath_all_err}
\end{table*}

\begin{table*}[tbh]
    \centering

\setlength{\tabcolsep}{5pt}
\begin{tabular}{L{2.cm} C{1.3cm} C{2.5cm} C{1.1cm} C{1.1cm}}
\toprule
    method & time [min] ($\downarrow$) & Loss ($\downarrow$) & memory [\%] ($\downarrow$) & params ($\downarrow$) \\
\midrule
    BitFit &            \underline{66} &                         0.19 $\pm$ 0.001738 &                \textbf{91} &     \underline{2.13M} \\
   DiffFit &               \textit{71} &       \textit{0.18} $\pm$ \textit{0.003334} &                \textit{93} &        \textit{2.58M} \\
     \ILoRA &                        77 &        \textbf{0.14} $\pm$ \textbf{0.00152} &             \underline{92} &     \underline{2.13M} \\
LoRA (QKV) &               \textbf{61} & \underline{0.17} $\pm$ \underline{0.005107} &             \underline{92} &        \textbf{1.23M} \\
\bottomrule
\end{tabular}
    \caption{Table for \Cref{fig:llama2_13b}}
    \label{tab:llama2_13b_to_metamath_all_err}
\end{table*}

\begin{table*}[tbh]
    \centering
    \setlength{\tabcolsep}{5pt}
\begin{tabular}{L{1.cm} C{1.3cm} C{1.45cm} C{1.1cm} C{1.1cm}}
\toprule
  method & time [h] ($\downarrow$) & FID ($\downarrow$) & memory [\%] ($\downarrow$) & params ($\downarrow$) \\
\midrule
MoRA$_1$ &            \textbf{9.6} &             13.9004 &                \textit{67} &               604.41K \\
MoRA$_6$ &                    23.5 &             14.0791 &                         87 &      \textbf{596.52K} \\
  BitFit &           \textit{13.0} &             13.8074 &             \underline{66} &   \underline{603.68K} \\
 DiffFit &                    14.3 &    \textit{13.6644} &                         87 &                 1.09M \\
    LoRA &                    19.5 &             13.7621 &                         79 &               864.03K \\
    VeRA &        \underline{12.0} & \underline{13.1129} &                         87 &      \textit{603.82K} \\
   \ILoRA &                    17.0 &    \textbf{12.8218} &                \textbf{65} &   \underline{603.68K} \\
\bottomrule
\end{tabular}
    \caption{Table for \Cref{fig:diffusion}}
    \label{tab:dit_to_food101_err}
\end{table*}

\begin{table*}[tbh]
    \centering

\setlength{\tabcolsep}{5pt}
\begin{tabular}{L{2.3cm} C{1.3cm} C{2.6cm} C{1.1cm} C{1.1cm}}
\toprule
       method & time [min] ($\downarrow$) & Accuracy ($\uparrow$) & memory [\%] ($\downarrow$) & params ($\downarrow$) \\
\midrule
     MoRA$_1$ &                        30 &        96.55 $\pm$ 0.0465&                 \textbf{6} &               333.36K \\
     MoRA$_6$ &                        31 &         97.5 $\pm$ 0.0544 &                 \textbf{6} &      \textbf{328.13K} \\
       BitFit &               \textbf{16} &      97.0 $\pm$ 0.055 &                 \textbf{6} &   \underline{331.78K} \\
      DiffFit &               \textit{18} &     \textit{97.9} $\pm$ \textit{0.0585}&                 \textit{8} &               817.15K \\
         LoRA &                        33 &                  \underline{98.1} $\pm$ \underline{0.058}&              \underline{7} &               663.55K \\
         VeRA &                        29 &                  97.7 $\pm$ 0.1631 &                 \textit{8} &      \textit{332.06K} \\
         DoRA &                        51 &                  \textbf{98.2} $\pm$ \textbf{0.0866} &                         11 &               995.33K \\
        \ILoRA &            \underline{17} &                  97.7 $\pm$ 0.186&                 \textbf{6} &   \underline{331.78K} \\
\ILoRA (norms) &               \textit{18} &                  \textit{98.0} $\pm$ \textit{0.1397} &                 \textbf{6} &               485.38K \\
\midrule
          All &                        22 &                  98.3 $\pm$ 0.0755&                         10 &               343.19M \\
\bottomrule
\end{tabular}
    \caption{Table for \Cref{fig:imagenet_to_cifar10}}
    \label{tab:pt_to_cifar10_all_err}
\end{table*}

\begin{table*}[tbh]
    \centering

\setlength{\tabcolsep}{5pt}
\begin{tabular}{L{2.2cm} C{1.3cm} C{2.5cm} C{1.1cm} C{1.1cm}}
\toprule
       method & time [min] ($\downarrow$) & Accuracy ($\uparrow$) & memory [\%] ($\downarrow$) & params ($\downarrow$) \\
\midrule
     MoRA$_1$ &                        31 & 82.7 $\pm$ 0.3253 &                 \textbf{6} &               333.36K \\
     MoRA$_6$ &                        30 &                 81.2 $\pm$ 0.1875 &                 \textbf{6} &      \textbf{328.13K} \\
       BitFit &               \textbf{16} &                 83.0 $\pm$ 0.204 &                 \textbf{6} &   \underline{331.78K} \\
      DiffFit &               \textit{18} &                          \textit{87.9} $\pm$ \textit{0.0775} &                 \textit{8} &               817.15K \\
         LoRA &                        34 &                          \textbf{88.4} $\pm$ \textbf{0.2173}&              \underline{7} &               663.55K \\
         VeRA &                        29 &                          85.3 $\pm$ 0.1723&                 \textit{8} &      \textit{332.06K} \\
         DoRA &                        50 &                          87.3 $\pm$ 0.2245&                         11 &               995.33K \\
        ILoRA &            \underline{17} &                          85.0 $\pm$ 0.567&                 \textbf{6} &   \underline{331.78K} \\
ILoRA (norms) &               \textit{18} &                          \underline{88.2} $\pm$ \underline{0.314} &                 \textbf{6} &               485.38K \\
\midrule
      All &                        22 &                          90.9 $\pm$ 0.3251&                         10 &               343.19M \\
\bottomrule
\end{tabular}
    \caption{Table for \Cref{fig:imagenet_to_cifar100}}
    \label{tab:pt_to_cifar100_all_err}
\end{table*}

\begin{table*}[tbh]
    \centering

\setlength{\tabcolsep}{5pt}
\begin{tabular}{L{3.1cm} C{1.3cm} C{2.8cm} C{1.1cm} C{1.1cm}}
\toprule
                          method & time [min] ($\downarrow$) & RMSE ($\downarrow$) & memory [\%] ($\downarrow$) & params ($\downarrow$) \\
\midrule
                           norms &                        65 &                        13.25 $\pm$ 0.004751 &                \textbf{36} &      \textbf{100.35K} \\
                $\beta$ (BitFit) &            \underline{61} &                        4.806 $\pm$ 0.003922 &             \underline{40} &   \underline{221.18K} \\
                        $\gamma$ &                        68 &                          4.17 $\pm$ 0.00482 &                \textit{41} &   \underline{221.18K} \\
                         DiffFit &                        66 &                         3.904 $\pm$ 0.00539 &             \underline{40} &      \textit{321.54K} \\
                           \ILoRA &               \textit{63} &        \textit{2.203} $\pm$ \textit{0.0078} &                \textbf{36} &   \underline{221.18K} \\
                 \ILoRA ($\beta$) &               \textbf{54} &      \textit{2.203} $\pm$ \textit{0.005498} &                \textbf{36} &               442.37K \\
                   \ILoRA (norms) &               \textbf{54} &        \textbf{2.19} $\pm$ \textbf{0.00655} &                \textbf{36} &      \textit{321.54K} \\
                \ILoRA ($\gamma$) &               \textit{63} &                       12.399 $\pm$ 0.006093 &                \textit{41} &               442.37K \\
          \ILoRA ($\beta$, norms) &                        67 & \underline{2.2} $\pm$ \underline{0.0051438} &                \textbf{36} &               542.72K \\
       \ILoRA ($\beta$, $\gamma$) &                        65 &                        4.166 $\pm$ 0.008069 &             \underline{40} &               663.55K \\
         \ILoRA ($\gamma$, norms) &                        73 &                        2.206 $\pm$ 0.001668 &                \textit{41} &               542.72K \\
\ILoRA ($\beta$, $\gamma$, norms) &                        75 &                        2.206 $\pm$ 0.002445 &             \underline{40} &               763.90K \\
\bottomrule
\end{tabular}
    \caption{Table for \Cref{fig:ablation_complementary_mde}}
    \label{tab:nyu_to_kitti_complementary_err}
\end{table*}

\begin{table*}[tbh]
    \centering

\setlength{\tabcolsep}{5pt}
\begin{tabular}{L{3.1cm} C{1.3cm} C{2.5cm} C{1.1cm} C{1.1cm}}
\toprule
                          method & time [min] ($\downarrow$) & Accuracy ($\uparrow$) & memory [\%] ($\downarrow$) & params ($\downarrow$) \\
\midrule
                           norms &               \textbf{16} &         97.7 $\pm$ 0.0395 &                 \textbf{6} &      \textbf{153.60K} \\
                $\beta$ (BitFit) &               \textbf{16} &         97.0 $\pm$ 0.055 &                 \textbf{6} &   \underline{331.78K} \\
                        $\gamma$ &            \underline{17} &      97.3 $\pm$ 0.0457&              \underline{8} &   \underline{331.78K} \\
                         DiffFit &               \textit{18} &     \underline{97.9} $\pm$ \underline{0.0585} &              \underline{8} &               817.15K \\
                           \ILoRA &            \underline{17} &         97.7 $\pm$ 0.186&                 \textbf{6} &   \underline{331.78K} \\
                 \ILoRA ($\beta$) &               \textit{18} &                  \textit{97.8} $\pm$ \textit{0.1245} &                 \textbf{6} &               663.55K \\
                   \ILoRA (norms) &               \textit{18} &                  \textbf{98.0} $\pm$ \textbf{0.139} &                 \textbf{6} &      \textit{485.38K} \\
                \ILoRA ($\gamma$) &                        19 &                  \textit{97.8} $\pm$ \textit{0.1043}&              \underline{8} &               663.55K \\
          \ILoRA ($\beta$, norms) &                        19 &     \underline{97.9} $\pm$ \underline{0.078} &                 \textbf{6} &               817.15K \\
       \ILoRA ($\beta$, $\gamma$) &                        21 &                  \textit{97.8} $\pm$ \textit{0.1297} &                 \textbf{6} &               995.33K \\
         \ILoRA ($\gamma$, norms) &                        20 &                  \textbf{98.0} $\pm$ \textbf{0.0837} &              \underline{8} &               817.15K \\
\ILoRA ($\beta$, $\gamma$, norms) &                        22 &                  \textbf{98.0} $\pm$ \textbf{0.109} &              \underline{8} &                 1.15M \\
\bottomrule
\end{tabular}
    \caption{Table for \Cref{fig:ablation_complementary_cls}}
    \label{tab:pt_to_cifar10_complementary_err}
\end{table*}

\end{document}